\documentclass[letterpaper]{article} % DO NOT CHANGE THIS
\usepackage[]{aaai23}  % DO NOT CHANGE THIS
\pdfoutput=1
\usepackage{times}  % DO NOT CHANGE THIS
\usepackage{helvet}  % DO NOT CHANGE THIS
\usepackage{courier}  % DO NOT CHANGE THIS
\usepackage[hyphens]{url}  % DO NOT CHANGE THIS
\usepackage{graphicx} % DO NOT CHANGE THIS
\usepackage{natbib}  % DO NOT CHANGE THIS AND DO NOT ADD ANY OPTIONS TO IT
\usepackage{caption} % DO NOT CHANGE THIS AND DO NOT ADD ANY OPTIONS TO IT
\usepackage{algorithm}
\usepackage{algorithmic}
\usepackage{newfloat}
\usepackage{listings}
\DeclareCaptionStyle{ruled}{labelfont=normalfont,labelsep=colon,strut=off} % DO NOT CHANGE THIS
\floatstyle{ruled}
\newfloat{listing}{tb}{lst}{}
\floatname{listing}{Listing}

\usepackage{booktabs}
\usepackage{graphicx}
\usepackage{subcaption}
\usepackage{natbib}
\usepackage{pdfpages}
\usepackage{amssymb,amsmath,graphicx}
\usepackage{pifont}% http://ctan.org/pkg/pifont
\usepackage{amsthm}
\usepackage{afterpage}
\usepackage{enumitem}
\usepackage{multirow}
\usepackage{microtype}

\newcommand{\myblue}[1]{#1}

\title{Neural Networks for Local Search and Crossover\linebreak in Vehicle Routing: A Possible Overkill?}

\author {
Ítalo Santana\textsuperscript{\rm 1},
Andrea Lodi\textsuperscript{\rm 2},
Thibaut Vidal\textsuperscript{\rm 1,3}
}
\affiliations {
	\textsuperscript{\rm 1} Department of Computer Science, Pontifical Catholic University of Rio de Janeiro\\
	\textsuperscript{\rm 2} Jacobs Technion-Cornell Institute, Cornell Tech and Technion - IIT\\
	\textsuperscript{\rm 3} CIRRELT \& SCALE-AI Chair in Data-Driven Supply Chains, Department of Mathematical and Industrial Engineering, Polytechnique Montr\'eal\\	
	isantana@inf.puc-rio.br, andrea.lodi@cornell.edu, thibaut.vidal@polymtl.ca
}

\begin{document}

\maketitle

\begin{abstract}
Extensive research has been conducted, over recent years, on various ways of enhancing heuristic search for combinatorial optimization problems with machine learning algorithms.
In this study, we investigate the use of predictions from graph neural networks (GNNs) in the form of heatmaps to improve the Hybrid Genetic Search (HGS), a state-of-the-art algorithm for the Capacitated Vehicle Routing Problem (CVRP).
The crossover and local-search components of HGS are instrumental in finding improved solutions, yet these components essentially rely on simple greedy or random choices. It seems intuitive to attempt to incorporate additional knowledge at these levels.
Throughout a vast experimental campaign on more than 10,000 problem instances, we show that exploiting more sophisticated strategies using measures of node relatedness (heatmaps, or simply distance) within these algorithmic components can significantly enhance performance. However, contrary to initial expectations, we also observed that heatmaps did not present significant advantages over simpler distance measures for these purposes. Therefore, we faced a common ---though rarely documented--- situation of overkill: GNNs can indeed improve performance on an important optimization task, but an ablation analysis demonstrated that simpler alternatives perform equally well.
\end{abstract}

\section{Introduction}
\label{sec:introduction}

Vehicle routing problems (VRP) represent one of the most studied classes of NP-hard problems due to their practical difficulty and ubiquity in real-life applications such as food distribution, parcel delivery, or waste collection, among others \citep{Toth2014,Vidal2020}.
Problems in this class generally seek to plan efficient itineraries for a fleet of vehicles to service a geographically-dispersed set of customers. The capacitated VRP (CVRP) is the most canonical variant among all existing routing problems. Its objective is to minimize the total distance traveled by the vehicles to service the customers, subject to a single constraint representing the vehicle capacities: i.e., the sum of customers' demands over a route should not exceed the vehicle capacity.

Over the years, there have been dramatic improvements in the heuristic and exact (i.e., provably optimal) solution of VRPs. To date, the best performing exact algorithms rely on branch-cut-and-price strategies, with tailored cutting-plane algorithms and sophisticated column-generation routines \citep{Costa2019, Pessoa2020}. With these methods, it is now possible to solve most existing instances with 200 or 300 customers. However, the time for an exact solution remains highly volatile at this scale, and most larger instances remain unsolved. Consequently, extensive research has been conducted on metaheuristics for this problem to find high-quality solutions in a shorter and more controlled time \citep{Vidal2013a}.

As it stands now, metaheuristics can consistently locate high-quality solutions for CVRPs with up to 1,000 customers in a matter of minutes~\citep{Christiaens2020,Vidal2022}. Of all existing methods, the Hybrid Genetic Search (HGS) algorithm developed in \citet{Vidal2012, Vidal2012b} and \cite{Vidal2022} is known to achieve the best-known solution quality consistently on most problems and instances of interest. Notably, during the 12th DIMACS implementation challenge on the CVRP organized in 2022 \citep{Dimacs2022}, it was used as the base algorithm for four out of the five best methods. Very-large problem instances counting dozens of thousands of customers can also be solved using tailored data structures and decomposition strategies~\citep{Accorsi2021, Santini2021}. Considering the latest generation of metaheuristics such as HGS, it is clear that two main operators ---local search and crossover--- are instrumental in finding improving solutions.
\begin{itemize}
\item \textbf{Local Search (LS)} consists in systematically exploring a neighborhood obtained by small changes over a current solution to identify improvements. This process is iterated until attaining a local minimum. Classical neighborhoods for the CVRP involve exchanges or relocations of client visits and edge reconnections. They typically include $\mathcal{O}(n^2)$ possible neighbors, where $n$ represents the number of customers. Due to its iterative nature, LS typically takes the largest share of the computational time. Several techniques have been developed to reduce computational complexity. In particular, \citet{Toth2003} observed that the search could be limited to relocations and exchanges of customers that are \emph{geographically related}. The resulting strategy, called granular search, limits classical neighborhoods to $\mathcal{O}(\Gamma n)$ moves, where $\Gamma$ is a user-defined parameter. However, although very simple in design, a straightforward distance-based relatedness criterion may hinder the search process, especially if optimal solutions require a few long edges.
\item In contrast, \textbf{Crossover} operators focus on diversifying the search. They consist of recombining two existing (parent) solutions into a new (offspring) solution that inherits promising characteristics from both. For the CVRP, crossover operators are not primarily designed for solution improvement, but instead used to create promising starting points for subsequent LS. Various crossovers have been used in previous works \citep{Nagata2007, Vidal2022}. As shown in Section~\ref{sec:methodology}, the Ordered Crossover (OX) is widely used, and consists in juxtaposing a fragment of the first solution with the remaining client visits ordered as in the second solution. By doing so, it implicitly creates a re-connection point, which is typically random.
\end{itemize}
Note that in both LS and Crossover, there is interest in using relatedness information between client vertices to (i) speed up the LS or (ii) identify a subset of more promising crossover operations. It is also noteworthy that the relatedness information used until now for the LS (and possibly used for the crossover) is a broader concept that goes beyond simple distance criteria, and which could be possibly learned.

In recent years, graph neural networks (GNNs) have emerged as a tool to apply machine learning techniques to combinatorial optimization problems posed over graphs. To the best of our knowledge, the first attempt in this context was proposed by \citet{Hopfield1985} for the TSP. Underpinned by enhancements in hardware and artificial intelligence research over the last years, the development of deep NNs made them relevant to a wide range of difficult combinatorial optimization problems, such as SAT, Minimum Vertex Cover, and Maximum Cut \citep{Dai2017, Yolcu2019}. When applied to solve CVRPs, these networks are usually combined with reinforcement learning~\citep[RL --][]{Nazari2018, Chen2019} or typically used for node classification or edge prediction~\citep{Kool2021, Xin2021b}.
Despite extensive research, GNNs for directly solving CVRPs remain limited to small problem instances with up to 100 customers and generally do not compare favorably with classic optimization methods (exact or heuristic) in terms of solution quality. This is possibly due to the fact that good solutions for combinatorial optimization problems result from tacit structural knowledge about the problem (learnable solution structure) along with a significant amount of trial-and-error to build the best possible solution fitting almost perfectly the constraints at hand. After all, an optimal solution is a very specific outlier. Whereas better knowledge of solution structures can be learned to guide the search, avoiding some (explicit or implicit) enumeration of solutions without compromising solution quality is generally challenging. Consequently, using pure learning algorithms without any other form of solution enumeration is likely to be unsuccessful.

Given these observations, a promising path toward better solution methods for VRPs concern the hybridization of learning-based and traditional solution methods. In this work, in particular, we aim to learn and use relatedness information in the LS and crossover operators of HGS to improve this state-of-the-art method substantially.
We capitalize upon the work of \citet{Kool2021}, which trained a GNN to predict occurrence probabilities of edges in high-quality solutions (i.e., heatmap), and used this information to sparsify the underlying graph and accelerate related solution procedures.
Instead, we leverage the heatmaps as a source of relatedness information to define neighborhood restrictions in the LS and possible re-connection points in the crossover. Throughout an extensive experimental campaign, we evaluate how HGS' performance varies with these surgical changes, for better or worse, on more than 10000 different instances containing from 100 to 1000 customers. Therefore, we make the following specific contributions.
\begin{enumerate}[nosep]
\item We introduce a framework for defining and exploiting \emph{relatedness} information between pairs of customers in the context of the CVRP.
\item We show that relatedness measures can be exploited to steer the LS towards the most promising moves. Additionally, we use relatedness information to extend the classical OX crossover, trading some of its inherent randomness for better choices of re-connection points between the parents. Our approaches are generic and applicable to any type of relatedness measure. We will specifically consider relatedness from two sources: geographical relatedness as given by the distance between two customers, and learnable relatedness (i.e., heatmap) obtained from a GNN.
\item We suggest a practical technique to exploit the output of a single GNN (heatmaps for fixed-size graphs) for problem instances of varying \myblue{sizes}. To achieve this, we decompose the original instance into a sequence of fixed-size subproblems and aggregate the resulting heatmap information. This approach has the benefit of only requiring a single trained model of moderate size.
\item Finally, we conduct an extensive computational campaign to measure the enhancements achieved with the proposed techniques and analyze the impact of each change. We observe that incorporating relatedness information within the crossover and LS operators largely benefit the search, such that learning-based approaches seem to be successful at first sight. However, after a closer analysis, we also observe that these improvements are mostly insensitive to the source of the relatedness information (geographical or learned). Therefore, the problem-specific knowledge and strategies that we integrated contributed more to the algorithm's performance than GNN-based algorithms for defining relatedness.
\end{enumerate}

\section{Methodology}
\label{sec:methodology}

The CVRP is defined over a complete graph $G=(V,E)$, where the set of vertices $V = \{0,1\dots,n\}$ contains a vertex~$0$ representing the depot, and the remaining vertices represent customers. Each customer $i \in \{1,\dots,n\}$ is characterized by a demand $d_i$. Edges $(i,j)$ model direct travel between vertices $i$ and $j$ for a distance $d_{ij}$. A solution to this problem is a set of routes originating and ending at the depot and visiting customers, such that (i) the total demand over each route does not exceed a vehicle-capacity limit $Q$, (ii) each customer is visited exactly once, and (iii)  the total travel distance is minimized. 

We additionally assume that we can calculate a \emph{relatedness} metric $\phi(i,j)$ for each edge $(i,j)$. This definition is general: in the simplest setting, relatedness could be the inverse of distance, i.e., $\phi(i,j) = 1/d_{ij}$. In a more informed setting, we can instead consider defining $\phi(i,j)$ as the output of a graph neural network (GNN) as seen in \citet{Kool2021}, predicting the probability of occurrence of an edge in a high-quality solution. Probabilities of this kind are typically called heatmaps. In the remainder of the paper, we will refer to $\phi_\textsc{d}(i,j)$ for distance-relatedness, and $\phi_\textsc{n}(i,j)$ for GNN-based relatedness. This information will now be used to refine the two most important HGS operators.

\subsection{Hybrid Genetic Search}
\label{subsec:hgs}

The Hybrid Genetic Search \cite{Vidal2022} relies on simple solution generation and improvement steps. A complete pseudo-code is provided in the \myblue{appendix}. The method starts by initializing a population of size $\mu$ with random solutions that are improved by local search. After this initialization phase, HGS iteratively generates new solutions by selecting two random solutions in the population, recombining them using an ordered crossover (OX), and applying local search for improvement. To promote exploration, solutions that exceed capacity limits are not directly rejected but instead penalized according to their amount of infeasibility. The penalty weights are adapted during the search to achieve a target percentage of feasible solutions, and infeasible solutions are maintained in a separate subpopulation. Whenever a solution is infeasible after local search, an extra \textsc{Repair} step is applied, which simply consists of a classic local search with a temporarily ($10 \times$) higher penalty coefficient.

During the overall search process, the number of solutions in the feasible and infeasible populations is monitored. Whenever any population exceeds $\mu + \lambda$ solutions, a survivors' selection phase is triggered to retain only the best $\mu$ individuals, according to a ranking metric based on solution value and contribution to the population diversity. Finally, the algorithm restarts each time $n_\textsc{it}$ consecutive solution generations have been done without improvement of the best solution, and it terminates upon a time limit $T_\textsc{max}$ by returning the best solution found over all the restarts.

\subsection{Local Search using Relatedness Measures} 
\label{subsec:neural-granular-search}

Local Search (LS) is a conceptually simple and efficient method to solve combinatorial optimization problems of the form $\min_{x \in X} c(x)$, where $X$ is the space of all solutions and~$c$ is the objective function. A neighborhood is defined as a mapping $\mathcal{N}: X \rightarrow 2^X$ associating with any solution~$x$ a set of neighbors $\mathcal{N}(x) \subset X$. For the CVRP, $\mathcal{N}(x)$ is usually defined relative to a set of operations (i.e., moves) that can modify the current solution $x$. A move $\tau$ is a small modification that can be applied on $x$ to obtain a neighbor $\tau(x) \in \mathcal{N}(x)$. HGS uses four main types of moves and some of their immediate extensions \citep{Vidal2022}:
\begin{itemize}[nosep]
    \item \textsc{Relocate}: Moves a visit to customer $i$ immediately after a visit to a different customer $j$ or the depot;
    \item \textsc{Swap}: Exchanges the visits of customers $i$ and $j$;
    \item \textsc{2-Opt}: Reverts a customer-visit sequence $(i,\dots,j)$;
    \item \textsc{2-Opt*}: Exchanges customers $i$ and $j$ and their succeeding visits.
\end{itemize}
The moves are evaluated in a random order of the indices $i$ and $j$, and any improvement is directly applied. This process is repeated until a local minimum is reached, i.e., a situation where no improving move exists for all the considered neighborhoods. Without further pruning, all these neighborhoods contain $O(n^2)$ solutions. Using incremental calculations (keeping track of partial load and distance over the routes), it is possible to conduct a complete evaluation of all neighborhoods in $O(n^2)$ time. Moreover, the number of complete neighborhood searches (i.e., loops) needed to converge is rarely greater than $10$ in practice.

\begin{figure}[t!]

\centering
\begin{minipage}{0.4\textwidth}
Twenty customers most related to customer $63$ according to $\phi_\textsc{d}$ and $\phi_\textsc{n}$:
\begin{center}
\boxed{\includegraphics[width=0.9\textwidth]{Figures/instance-distance-heatmap.pdf}}
\end{center}
\end{minipage}

\vspace*{0.4cm}

\begin{minipage}{0.4\textwidth}
\begin{center}
Customer $63$ in an optimal solution:
\begin{center}
\boxed{\includegraphics[width=0.5\textwidth]{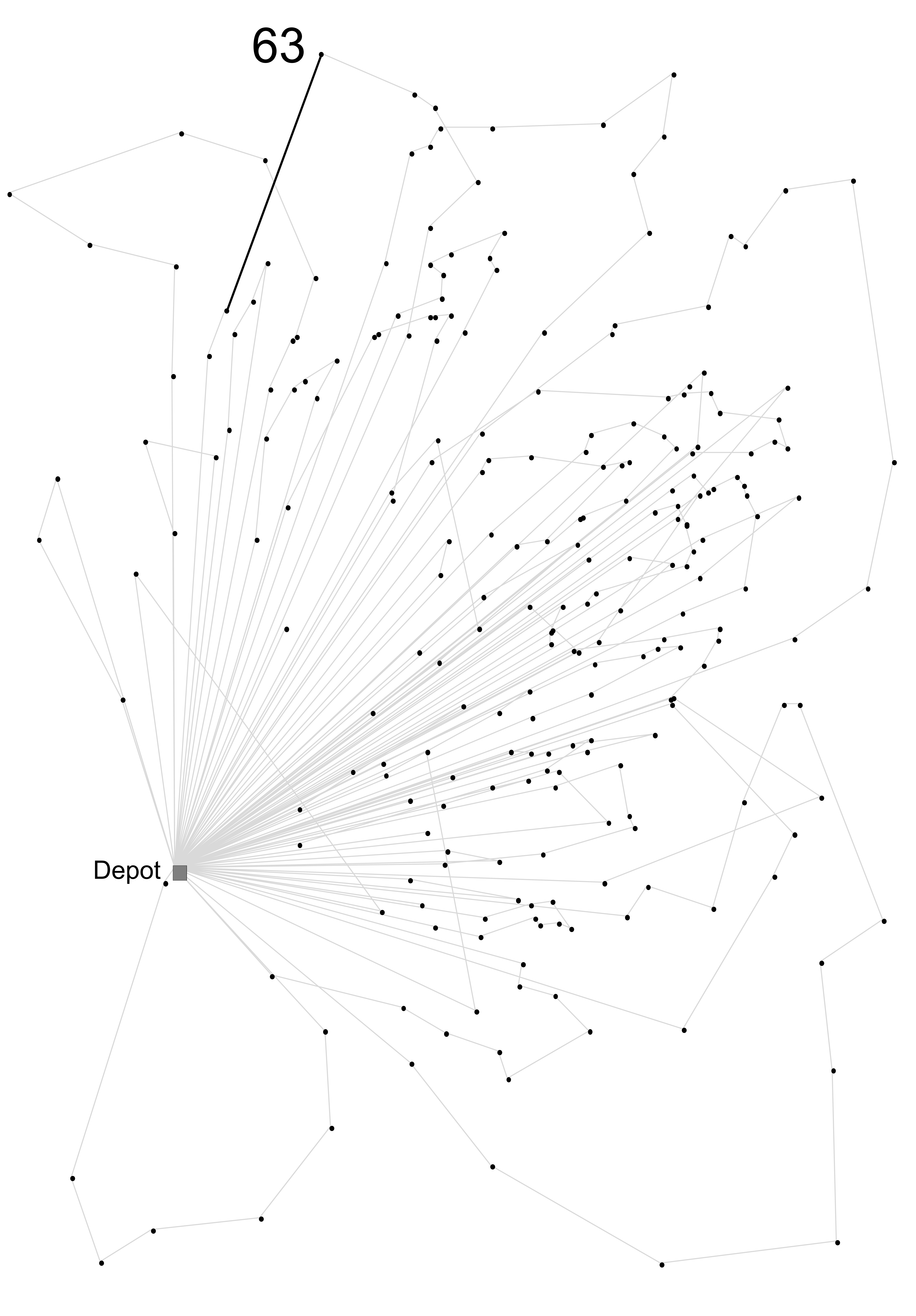}}
\end{center}
\end{center}
\end{minipage}
\caption{Sets of related customers according to $\phi_\textsc{d}$ and $\phi_\textsc{n}$, on instance X-n247-k50}%\vspace*{0.3cm} %% This command is not recommended in the authors' kit
\label{fig:relatedness-illustration}
\end{figure}

A quadratic complexity for the LS operator is adequate for small problems, but this can become a significant bottleneck otherwise due to its frequent use. Considering this, \citet{Toth2003} introduced a ``granular search'' mechanism that consists in limiting the moves to customer pairs $(i,j)$ that are geographically close, i.e., such that $j$ belongs to a set $\Phi(i)$ formed of the $\Gamma$ closest customers of $i$. Consequently, the total number of moves and the complexity of each LS loop reduce down to $O(n \Gamma)$ time. Indeed, it rarely makes sense to relocate or exchange customer visits that are far away from each other. Moreover, this strategy ensures that each move creates at least one short edge  \citep{Schneider2017,Vidal2013,Accorsi2021}. 

Since its inception, granular search has been adapted to many VRP variants. Especially, to handle customer constraints on service-time windows, \cite{Vidal2013} extended the concept to filter node pairs $(i,j)$ based on a compound metric that includes distance, unavoidable waiting times, and unavoidable time-window violations arising from this customer succession. 

In this study, we instead extend the filtering criterion by relying on relatedness information from the GNN. As illustrated in Figure~\ref{fig:relatedness-illustration} for instance X-n247-k50 from \citet{Uchoa2017}, the $\Gamma$ most-related customers according to the relatedness metrics $\phi_\textsc{d}$ and $\phi_\textsc{n}$ can differ very significantly. In this particular example, the GNN-based relatedness even includes an edge (in boldface) contained in the optimal solution that is otherwise missing when considering distance only.

For each $i$, we therefore form the set $\Phi(i)$ in two steps: we first include in $\Phi(i)$ the $\lfloor \Gamma/2 \rfloor$ vertices that are most related to $i$ according to the GNN-based relatedness metric $\phi_\textsc{n}(i,j)$, and then we complete the $\lceil \Gamma/2 \rceil$ remaining customers by increasing distance, therefore according to $\phi_\textsc{d}(i,j)$. This strategy uses learned information and ensures that the $\Gamma/2$ closest customers are still considered in the moves.

\subsection{Crossover using Relatedness Measures}
\label{subsec:neural-ox-crossover}

In HGS, each solution is represented as a single permutation of the customer's visits (i.e., a giant tour) during the crossover operation. This use of this simple representation is motivated by the fact that (i) one can simply represent any complete solution by concatenating the routes and omitting the visits to the depot, and (ii) reversely, given a sequence of customers visits, there exists a linear-time algorithm, called \textsc{Split}, that optimally segments this giant tour into routes \citep{Vidal2016}. 

\begin{figure}[t]
\centering
\centerline{\includegraphics[width=\columnwidth]{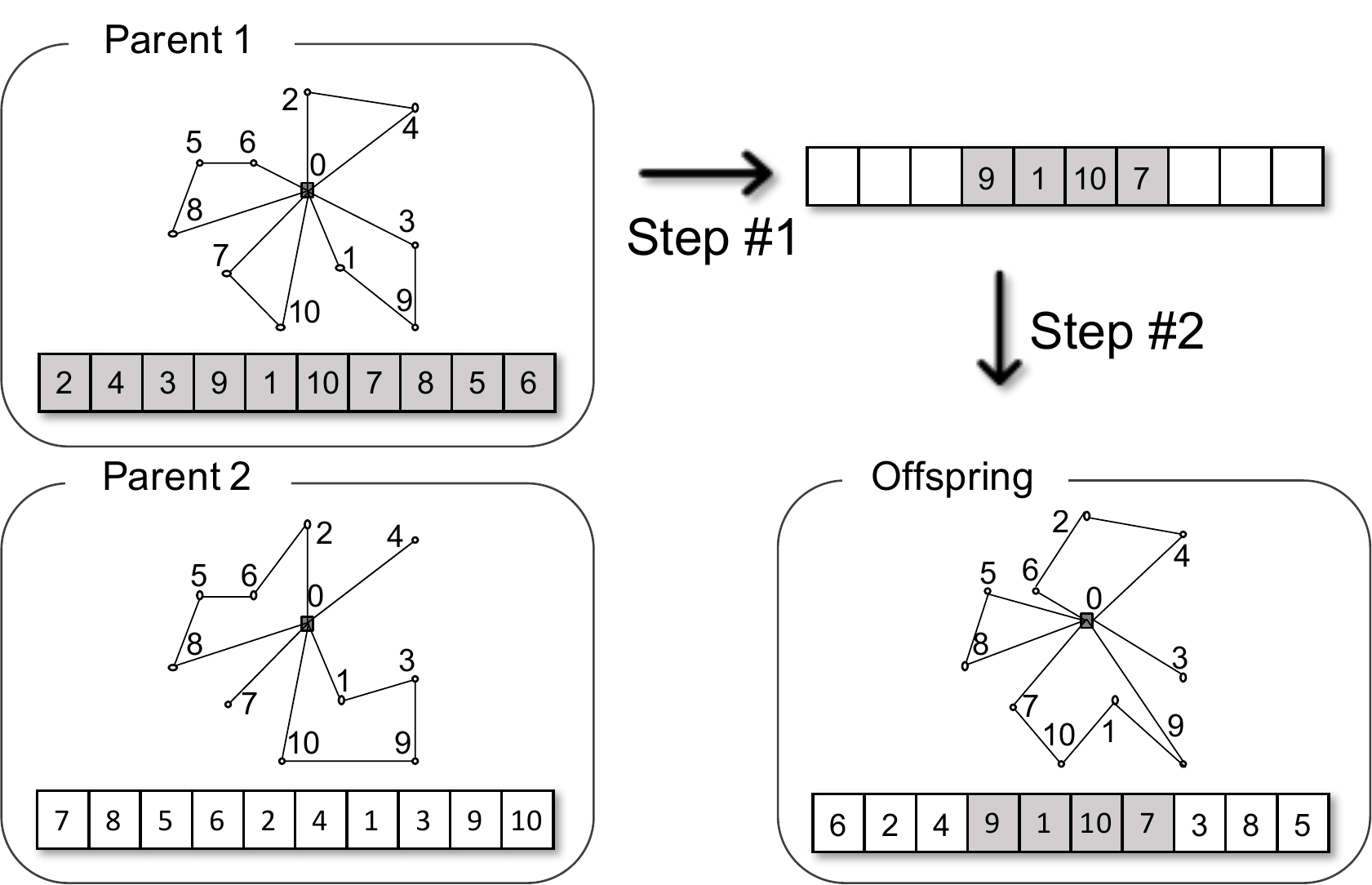}}
\caption{Illustration of the ordered crossover (OX)}
\label{fig:ox-crossover}
\end{figure}

Based on this representation, HGS employs the ordered crossover (OX -- \citealt{Oliver1987}) illustrated in Figure~\ref{fig:ox-crossover}.
OX works in two steps. First, a fragment $F$ of the first parent defined by two randomly-selected cutting points is copied in place into an empty offspring. Next, the second parent is scanned from the position of the second cutting point to complete all missing customer visits circularly. This gives a new giant tour, which is then transformed into a complete CVRP solution using \textsc{Split}.

As it stands, OX is completely dependent upon random choices. In particular, the second step tends to concatenate unrelated customers immediately after fragment~$F$. This creates low-quality fragments of solution requiring many LS moves for improvement. To correct this issue, we suggest relying on the relatedness metric to modify the completion step. Let $i$ be the last customer from fragment $F$. Instead of arbitrarily reconnecting $F$ with the next customer from Parent~2 obtained by a circular sweep, we select a random related customer $j$ among the $\Gamma$ customers most related to $i$ that are not part of $F$ \myblue{(or a random position if no such $j$ exists)} and then proceed to complete the offspring from this position following the order in Parent~2. This small but notable difference permits reconnecting visits that are more closely related among both parents, allowing for better solutions without sacrificing diversity. As previously, the choice of relatedness metric leads to different variants of the OX crossover. An illustrative example of such re-connection is presented in the \myblue{appendix}. In the remainder of this paper, we will refer to the modified crossover using distance-relatedness as DOX, and to the modified crossover using GNN-relatedness as NOX.

\section{Experimental Analyses}
\label{sec:experiments}

This section presents extensive computational experiments designed to:
(i) calibrate and evaluate the impact of the granular search parameter $\Gamma$, which governs the size of the LS neighborhoods;
(ii) measure the impact of our enhancements on the LS and OX operators as well as the usefulness of different relatedness criteria;
(iii) confront the characteristics, the computational effort, and the performance of heatmaps produced by different GNN configurations, and
(iv) analyze the extension of GNNs originally trained on fixed-size graphs to instances of varying sizes.
We address objective (i) in Section~\ref{subsec:results-gamma-calibration}, whereas objectives (ii, iii, iv) are covered in Sections~\ref{subsec:results-XML-instances}~and~\ref{subsec:results-X-instances}.\\

We will analyze, in the following sections, the performance of HGS in its original form (baseline) along with five combinations of $\phi_\textsc{d}$ and $\phi_\textsc{n}$ for local search and crossover operators, which are listed below:
\begin{itemize}[nosep]
\item \textbf{HGS-D-O (baseline)}: HGS with granular search and OX;
\item \textbf{HGS-D-D}: HGS with granular search and DOX;
\item \textbf{HGS-D-N}: HGS with granular search and NOX;
\item \textbf{HGS-N-O}: HGS with neural granular search and OX;
\item \textbf{HGS-N-D}: HGS with neural granular search and DOX;
\item \textbf{HGS-N-N}: HGS with neural granular search and NOX.
\end{itemize}

\subsection{Computational Environment}

All experiments are run on a single thread of an Intel Gold 6148 Skylake 2.4\,GHz processor with 40\,GB of RAM and NVIDIA Tesla P100 Pascal (12\,G memory), running CentOS\,7.8.2003. Unless otherwise stated, we use the original parameters defined for HGS in~\citet{Vidal2022} and the GNN in~\citet{Kool2021}. To achieve fast convergence, we set smaller values for the population-size parameters in HGS: $\mu = 12$ and $\lambda = 20$. We compile HGS with g++\,9.1.0 and execute the GNN using Python\,3.8.8 on Torch\,1.9.1.

\subsection{Benchmark Instances}

Our experiments use two main sets of CVRP instances: Set X from~\citet{Uchoa2017}, and Set XML from~\citet{Queiroga2022}.
Set~X is a well-known benchmark of 100 instances containing between 100 and 1000 customers. This set includes very diverse instances that mimic important characteristics of real-world situations concerning depot positions, route length, customer demands, and locations.
The XML set~\citep{Queiroga2022} includes 10,000 instances of 100 customers each, drawn from a similar distribution as set X. One advantage of the XML set is that \myblue{the} number of customers is constant in all instances, and all optimal solutions are provided. This permits comparisons with proven optima instead of best-known solutions (BKS) collected from all previous works. In contrast, many instances of set X are still unsolved to proven optimality. All instance sets, open-source codes, and scripts needed to run the experiments are provided at \myblue{\url{[released-upon-final-publication]}}.

\subsection{Parametrization and Training of the GNN}

The GNN proposed by \citet{Kool2021} is designed to be trained and applied for prediction over graphs (i.e., instances) of fixed size. The authors provided the final model trained on instances containing 100 customers, which can therefore be directly applied for inference on the XML instances. In contrast, Set X has instances with different numbers of customers, such that a different approach is needed to use the heatmaps. Due to these key differences, we will subdivide the presentation of our experiments into two parts, with results on XML instances in Section~\ref{subsec:results-XML-instances}, and adaptations and results for Set X in Section~\ref{subsec:results-X-instances}.

In these experiments, we use the original trained GNN from~\citet{Kool2021} for heatmap generation, called~\textsc{Original} in the rest of this paper. However, even though this model is already trained, it takes around $0.85$\,seconds of inference time to produce the heatmap for a given instance. This is a similar order of magnitude as the time needed by HGS to solve the CVRP to near optimality (i.e., below $0.1\%$ error) on instances containing 100 customers. Since we aim to compare CVRP solution algorithms under the same total CPU time budget (counting inference time and solution time), GNN-based methods would be at a disadvantage if a large share of the CPU time is invested in the inference step. Therefore, to estimate the performance of GNN-based algorithms in the most optimistic conditions (e.g., considering a hypothetical scenario where GPU inference is extremely fast), we will also report the results of the same method ignoring inference time. Additionally, we produce results (counting inference time) obtained with two lighter versions of the GNN, called \textsc{Model \#1} and \textsc{Model \#2}, which were trained on the same examples as \citet{Kool2021}, with fewer internal nodes and internal layers. Table~\ref{tab:model-configurations} summarizes the parameter setting of all the considered GNNs.

\begin{table}[ht]
\caption{GNN configurations}
\label{tab:model-configurations}
\resizebox{\columnwidth}{!}
{
\centering
\begin{sc}
\begin{tabular}{crrrr}
\toprule
GNN & \#Nodes & \#Layers & \#Epochs & Pred-T(s)\\
\midrule
Original & 300 & 30 & 1500 & 0.85\\
Optimistic & 300 & 30 & 1500 & Ignored\\
Model \#1 & 10 & 5 & 500 & 0.03\\
Model \#2 & 10 & 5 & 1500 & 0.03\\
\bottomrule
\end{tabular}
\end{sc}
}
\end{table}

This table lists for each GNN the number of hidden layers (\# \textsc{Layers}), nodes per layer (\textsc{\# Nodes}), and epochs (\textsc{\# Epochs}) used for training. Finally, the last column reports the average inference time on an XML instance. The parameters of \textsc{Model \#1} and \textsc{Model \#2} were selected to achieve training and inference in a limited time. \textsc{Model \#1} (resp. \textsc{\#2}) required 8 (resp. 24) hours of training time on our hardware.

\subsection{Calibration of the Local Search}
\label{subsec:results-gamma-calibration}

We focus here on the parameter $\Gamma$, which drives the exploration breadth of the LS (see~Section~\ref{subsec:neural-granular-search}) and significantly impacts the computational time of HGS. The aim of this experiment is to select a meaningful range of values for this parameter. Based on standard values used in previous works, we evaluate configurations $\Gamma \in \{5,10,15,20,30,50,100\}$ and analyze the sensitivity of the baseline method (i.e., HGS-D-O) to this parameter. To keep a simple experimental design, we focus on the performance of the LS by generating ten random initial solutions for each instance and applying a single LS to each of these solutions. We then report in Table~\ref{tab:granular-search} the quality of the best solution found as well as the computational time used by the ten LS runs.

\begin{table}[htbp]
\caption{Impact of $\Gamma$ on solution quality and CPU time}
\label{tab:granular-search}
\centering
\begin{sc}
\begin{tabular}{ccccccc}
\toprule
 & \multicolumn{2}{c}{Set X} & \multicolumn{2}{c}{Set XML}\\\
 $\Gamma$ & Gap\% & Time (s)& Gap\% & Time (s)\\
\midrule
5	&  4.664 & 0.157 & 2.976 & 0.024  \\ 
10	&  4.018 & 0.160 & 2.365 & 0.026\\ 
15 &   3.817 & 0.162 & 2.194 & 0.027\\
20	&  3.667 & 0.165 & 2.137 & 0.029\\ 
30	&  3.630 & 0.197 & 2.087 & 0.034\\ 
50	&  3.696 & 0.238 & 2.089 & 0.043\\ 
100	&  3.690 & 0.375 & 2.087 & 0.057 \\ 
\bottomrule
\end{tabular}
\end{sc}
\end{table}

In Table \ref{tab:granular-search} and the rest of this paper, solution quality is expressed as a percentage error gap calculated as $\textsc{Gap(\%)} = 100 \times (z - z_\textsc{bks})/z_\textsc{bks}$, where $z$ represents the cost of the solution and $z_\textsc{bks}$ is the optimal or BKS cost value.

The results of this experiment indicate that solution quality generally improves with $\Gamma$, but with decreasing marginal returns. We cease to see notable solution quality improvements once $\Gamma$ exceeds a value of $30$, but CPU time dramatically increases. Given this, we set $\Gamma = 15$ in the remainder of our experiments, and additionally provide detailed results with $\Gamma \in \{20,30,50\}$ in the \myblue{appendix}.

\subsection{Experimental Results -- Set XML}
\label{subsec:results-XML-instances}

Having calibrated all the algorithmic components, we can now measure the impact of GNN-informed relatedness measures in the LS and crossover operator. We focus here on the instances of set XML. Given that HGS converges towards near-optimal solutions within seconds for these instances, \myblue{we use a short termination criterion with} $T_\textsc{max} = 5$\,seconds per instance and report final results as well as convergence plots to measure the impact of the different versions of the LS and crossover.

\begin{figure}[ht!]
\centering
\centerline{\includegraphics[width=0.85\columnwidth]{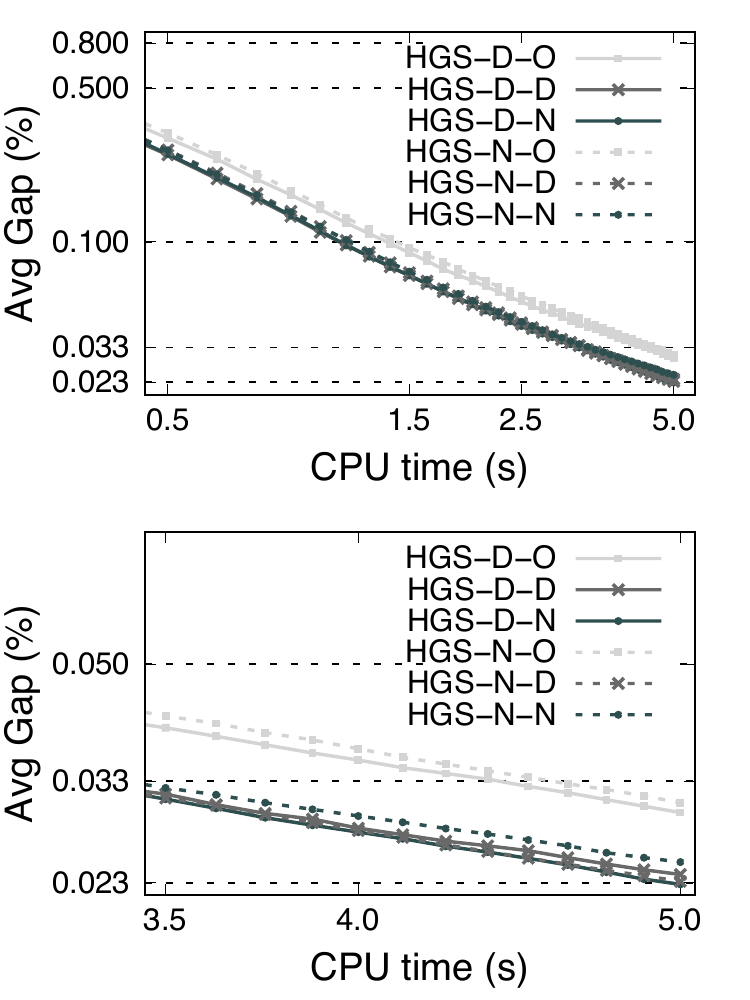}}
\caption{Convergence plots for all HGS variants on set XML (upper graph = complete run, lower graph = last 1.5\,seconds)}
\label{fig:convergence-plot-XML}
\end{figure}

Table~\ref{tab:results-XML} therefore reports the number of optimal solutions (\textsc{\#Opt}) attained over the 10,000 instances and the average final \textsc{Gap(\%)} for all of the methods, considering the four possible GNN configurations (\textsc{Original}, \textsc{Optimistic}, \textsc{Model \#1}, and \textsc{Model \#2}). Best performance is indicated in boldface.
Additionally, the convergence plots of Figure~\ref{fig:convergence-plot-XML} depict the progress of the average gap of the different methods over time for the \textsc{Optimistic} configuration of the GNN, and similar graphs are provided for the other GNN configurations in the \myblue{appendix}.

\begin{table*}[ht!]
\caption{Results of all methods and GNN configurations for the instances of set XML}
\label{tab:results-XML}
\centering
\resizebox{\textwidth}{!}
{
\begin{sc}
\begin{tabular}{crrrrrrrrrrrr}
\toprule
 & \multicolumn{2}{c}{HGS-D-O} & \multicolumn{2}{c}{HGS-D-D} & \multicolumn{2}{c}{HGS-D-N} & \multicolumn{2}{c}{HGS-N-O}  & \multicolumn{2}{c}{HGS-N-D}  & \multicolumn{2}{c}{HGS-N-N} \\
 GNN & \#Opt &  Gap\% &  \#Opt &  Gap\% &  \#Opt &  Gap\% &  \#Opt &  Gap\%&  \#Opt &  Gap\%&  \#Opt &  Gap\%\\
\midrule
Original  &7715 &0.030 &\textbf{8105} &0.024 &8086 &\textbf{0.023} &7691 &0.031 &8046 &\textbf{0.023} &8011 & 0.025\\
Optimistic  &7715 &0.030 &8105 &0.024 &\textbf{8120} &\textbf{0.023} &7732 &0.031 &8062 &\textbf{0.023} &8041 & 0.025\\
Model \#1  &7715 &0.030 &\textbf{8105} &0.024 &8094 &\textbf{0.023} &7697 &0.031 &8028 &0.024 &7994 & 0.025 \\
Model \#2 &7715 &0.030 &\textbf{8105} &\textbf{0.024}&8102 &\textbf{0.024} &7719 &0.030 &8067 &\textbf{0.024} &8057 & \textbf{0.024} \\
\bottomrule
\\
\end{tabular}
\vspace*{0.5cm} %% TV -- just to improve a bit the layout for the ArXiV report
\end{sc}
}
\end{table*} 

As seen in these results, all HGS versions progress \myblue{(i.e., decrease the gap) smoothly} within the time limit, and all approaches except \mbox{HGS-N-O} outperformed HGS-D-O (the baseline HGS algorithm) in terms of their number of optimal solutions and average gap. The significance of these improvements is confirmed by two-tailed paired-samples Wilcoxon tests between each \myblue{method} and HGS-D-O at a significance level of 0.05.

However, these experiments also show that \mbox{HGS-N-O} does not perform significantly better than \mbox{HGS-D-O}, even when ignoring the inference time (i.e., \textsc{Optimistic} evaluation of the GNN). This indicates that \myblue{using} the GNN-based relatedness criterion in the LS does not bring significant benefits. It is an open research question to determine if different GNN architectures may perform better in the task of filtering LS neighborhoods.

Now, a comparison of configurations HGS-D-O (baseline), HGS-D-D, and HGS-D-N permits us to assess the impact of our changes on the crossover operator. We remind that HGS-D-O refers to the original OX crossover, whereas HGS-D-D and HGS-D-N modify the reconnection step to integrate relatedness information. As seen in our experiments, HGS-D-D and HGS-D-N are much better than the baseline (final gaps of $0.024\%$ compared to $0.030\%$), as confirmed by paired-samples Wilcoxon tests at $0.05$ significance level. This is a notable breakthrough, given that it is uncommon to identify simple conceptual changes to HGS that significantly improve its state-of-the-art performance.

Finally, the choice of configuration for the GNN did not significantly affect the results, and our observations remain valid for the \textsc{Original}, \textsc{Model \#1}, and \textsc{Model \#2} configurations.

\subsection{Experimental Results -- Set X}
\label{subsec:results-X-instances}

\begin{table*}[t]
\caption{Results of all methods and GNN configurations for the instances of set X}
\label{tab:results-X}
\centering
\resizebox{\textwidth}{!}
{
\begin{sc}
\begin{tabular}{crrrrrrrrrrrr}
\toprule
  &\multicolumn{2}{c}{HGS-D-O} &  \multicolumn{2}{c}{HGS-D-D} & \multicolumn{2}{c}{HGS-D-N}  &  \multicolumn{2}{c}{HGS-N-O}  & \multicolumn{2}{c}{HGS-N-D}&  \multicolumn{2}{c}{HGS-N-N} \\
GNN &  \#Opt &   Gap\% &   \#Opt &   Gap\% &   \#Opt &   Gap\%&   \#Opt &   Gap\%&   \#Opt &   Gap\% &   \#Opt &   Gap\%\\
\midrule
Original & \textbf{188}& 0.368& 187& \textbf{0.302}& 184& 0.317& 177& 0.395& 169& 0.325& 172&  0.317\\
Optimistic & \textbf{188}& 0.368& 187& 0.302&187& \textbf{0.299}& 185& 0.365& 177& 0.307& 182&  \textbf{0.299}\\
Model \#1 & \textbf{188}& 0.368& 187& \textbf{0.302}&185& 0.309& 166& 0.414& 177& 0.332& 184&  0.336\\
Model \#2 & \textbf{188}& 0.368& 187& 0.302& 186& \textbf{0.301}& 169& 0.391& 175& 0.314& 170&  0.316\\
\bottomrule
\\
\end{tabular}
\vspace*{0.5cm} %% TV -- just to improve a bit the layout for the ArXiV report
\end{sc}
}
\end{table*} 

The instances of set X include a different number of customers, but the GNN of \citet{Kool2021} is designed to predict heatmaps only for fixed-size graphs. Moreover, training a model on an instance of maximal size (1000 customers) and relying on dummy nodes is likely to require extensive training time (especially with its original parametrization).

To circumvent this issue, we instead propose to combine the heatmaps from different subproblems to obtain a relatedness measure for all customers. Let $n_\textsc{g}$ be the graph size handled by the GNN. For each customer $i \in \{1,\dots,n\}$, in turn, we collect the $n_\textsc{g}-1$ closest customers along with the depot to form a CVRP subproblem with exactly $n_\textsc{g}$ customers. We rely on the GNN to infer the heatmap for this graph, and use the heatmap values for all edges $(i,j)$ such that $j$ belongs to the subproblem and $0$ otherwise. This simple approach requires $n$ heatmaps inference steps, but the inherent parallelism of Pytorch makes it effective enough for our purposes.

\begin{figure}[!ht]
\centering
\centerline{\includegraphics[width=0.85\columnwidth]{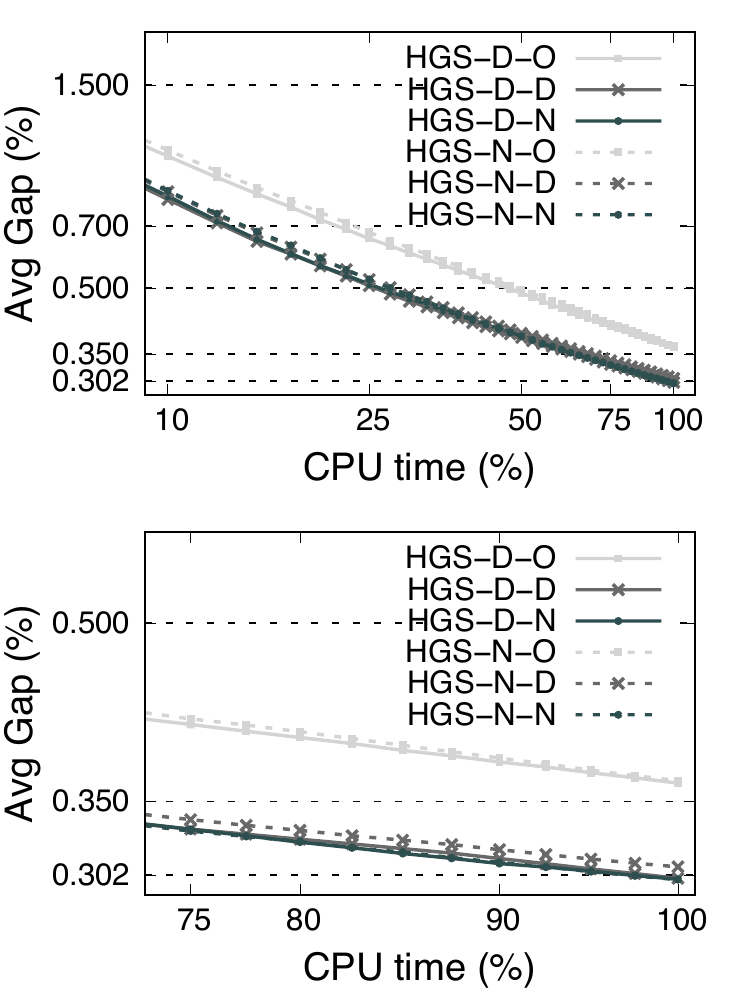}}
\caption{Convergence plots for all HGS variants on set X}
\label{fig:convergence-plot-X}
\end{figure}

As previously, we report the results of the different HGS variants in~Table~\ref{tab:results-X} for the four considered GNN parameter settings. We set a total computational time budget that is linearly proportional to $n$, allowing 24 seconds for the smallest instance (X-n101-k25) with 100 customers, and up to 240 seconds for the largest one (X-n1001-k43) with 1000 customers. Moreover, we perform $10$ experiments with different random seeds for each of the $100$ instances, leading to $1000$ solution processes. The table, therefore, counts the number of optimal solutions out of 1000 as well as the average error gap when the algorithm terminates. With these time limits, the inference time of the \textsc{Original} GNN represents 15.2\% of the overall time budget, and the inference time of \textsc{Model \#1} and \textsc{Model \#2} is limited to 1.9\% of the time budget. Convergence plots in the same format as before are additionally presented in~Figure~\ref{fig:convergence-plot-X}.

These additional results on Set X confirm our previous observations: all HGS variants except HGS-N-O outperformed the HGS-D-O baseline. Additionally, the proposed modifications to the crossover operator (HGS-D-D and HGS-D-N) led to performance improvements that are even more expressive on that instance set, with final gaps of $0.302\%$ and $0.299\%$ compared to $0.368\%$ for the original HGS. As previously, however, \myblue{using} learned information from the GNN instead of distance did not make a substantial difference in the crossover and even appeared to be detrimental in the context of the LS. 

It is important to stress that, without a complete analysis involving HGS-D-D, a comparison of HGS-D-N versus HGS-D-O could have led to the conclusion that the GNN was responsible for the improvement. However, recommending the use of this method in this context would have been an ``overkill'' since a simpler reconnection mechanism based on distance effectively produces the same gains.

\section{Conclusions}
\label{sec:conclusion}

In this work, we have shown that relatedness metrics can be broadly used to improve the performance of the HGS \citep{Vidal2022}, a state-of-the-art solution algorithm for the CVRP. Relatedness has been exploited in two ways: to focus the LS on promising moves, and to steer the crossover operator towards meaningful reconnections. As relatedness is a fairly general concept, we can freely use geographical or learnable (i.e., GNN-based) information for that purpose. As seen in our experimental analyses, these adaptations lead to significant improvements on a large benchmark counting over 10,000 instances. Additionally, we show that a simple strategy to extend GNN heatmap predictions to instances of varying size is fairly effective, circumventing the limitation due to fixed-size training. Overall, exploiting heatmaps to boost HGS operators is very effective, but also not superior to a simpler application of distance-based relatedness for similar purposes. This observation \myblue{contrasts} with the superiority claims of sophisticated learning mechanisms and ever-larger networks. Instead, it aligns with the ``less-is-more'' approach toward algorithmic design.

We acknowledge that some aspects studied in this work can be further investigated for future research. The first one refers to the applications of relatedness criteria to other combinatorial optimization \myblue{problems} and solvers (e.g., branch and bound). Another research avenue of interest concerns exploiting different relatedness sources and simpler machine learning models. Finally, from a more general viewpoint, we expect that the contributions of this work can lead to a better comprehension of the challenges involved in incorporating sophisticated machine learning techniques into state-of-the-art solvers. We believe that \myblue{research on} GNN-enhanced heuristics is promising, \myblue{but that} careful ablation studies are \myblue{essential to correctly measure impacts and improvements}.

\section*{Acknowledgements}
\myblue{This research was enabled in part by support provided by Calcul Québec, Compute Canada, and CAPES-PROEX [grant number 88887.214468/2018-00]. This support is gratefully acknowledged.}

\appendix
\newpage
\onecolumn

\section{Pseudo-code of the Hybrid Genetic Search (HGS)}

Algorithm~\ref{alg:HGS} provides the pseudo-code of HGS for the CVRP, as originally proposed in \citet{Vidal2022} and described in Section~2.1 of the main paper.

\begin{figure*}[hp!]
	\centering
	\begin{minipage}{0.65\textwidth}
		\begin{algorithm}[H]
			\caption{Hybrid Genetic Search for the CVRP (HGS)}
			\label{alg:HGS}
			\begin{algorithmic}[1]
				\STATE Initialize population with random solutions improved by local search
				\WHILE{$time < T_{max}$}
				\STATE Select parent solutions $P_1$ and $P_2$
				\STATE Apply the crossover operator on $P_1$ and $P_2$ to generate an offspring $C$
				\STATE Educate offspring $C$ by local search
				\STATE Insert $C$ into respective subpopulation
				\IF{$C$ is infeasible}
				\STATE With 50\% probability, repair $C$ (local search) and \\
				\STATE insert it into respective subpopulation
				\ENDIF
				\IF{maximum subpopulation size reached}
				\STATE Select survivors
				\ENDIF
				\STATE Adjust penalty coefficients for infeasibility
				\ENDWHILE
				\STATE Return best feasible solution
			\end{algorithmic}
		\end{algorithm}
	\end{minipage}
\end{figure*}

\section{Illustration of the NOX and DOX Operators}

\begin{figure*}[htbp]
	\centering
	\includegraphics[width=0.6\textwidth]{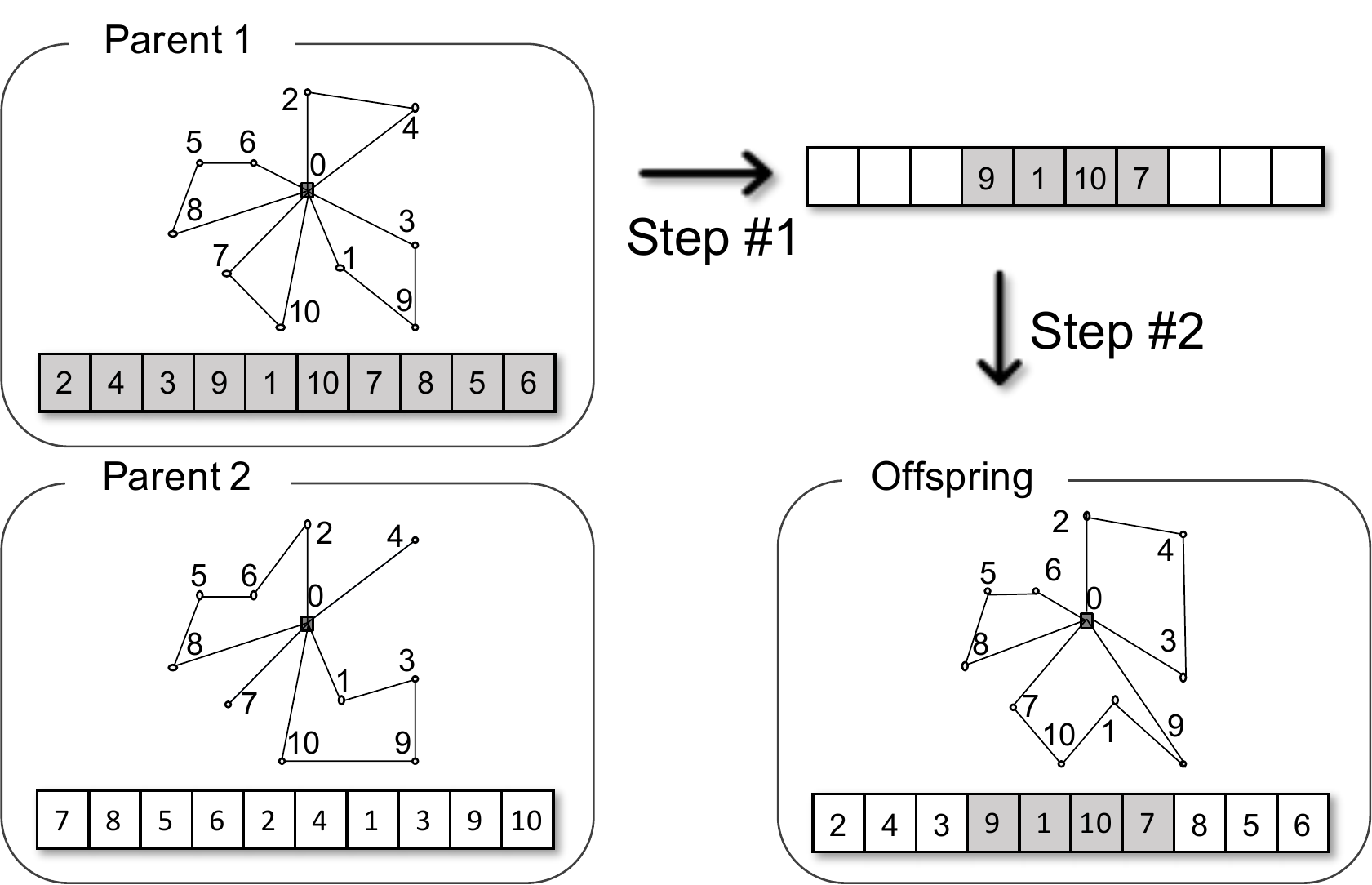}
	\caption{Illustration of the crossover using relatedness measures (NOX and DOX)}
	\label{fig:crossover-with-relatedness-measures}
\end{figure*}

Figure~\ref{fig:crossover-with-relatedness-measures} illustrates the use of relatedness measures within OX, as discussed in Section~2.3. In this example, which is valid for both NOX and DOX, the algorithm selects fragment $F = [9,1,10,7]$ from Parent 1. With this, $i=7$ is the last element of $F$ (Step \#1). Then, it proceeds to select a random related customer among the $\Gamma$ most-related customers of $i$ that are not part of~$F$. We assume, in this example, that this customer is $j = 8$. Finally, it completes the offspring from this position, following the order in Parent 2 (Step \#2).

\section{Detailed Experimental Results -- Set XML}

Table~\ref{tab:results-XML-appendix} provides additional detailed results, covering all values of $\Gamma \in \{15,20,30,50\}$ on all XML~instances of \citet{Queiroga2022}. It reports the number of optimal solutions (\textsc{\#Opt}) attained over the 10,000 XML~instances and the average final \textsc{Gap(\%)}, calculated as $\textsc{Gap(\%)} = 100 \times (z - z_\textsc{bks})/z_\textsc{bks}$, where $z$ represents the cost of the solution and $z_\textsc{bks}$ is the optimal or BKS cost value. The results are provided for all of the methods, considering the four possible GNN configurations (\textsc{Original}, \textsc{Optimistic}, \textsc{Model \#1}, and \textsc{Model \#2}). The best performance in each row is highlighted in boldface.

\begin{figure*}[hp!]
	\begin{center}
		\includegraphics[width=0.7\textwidth]{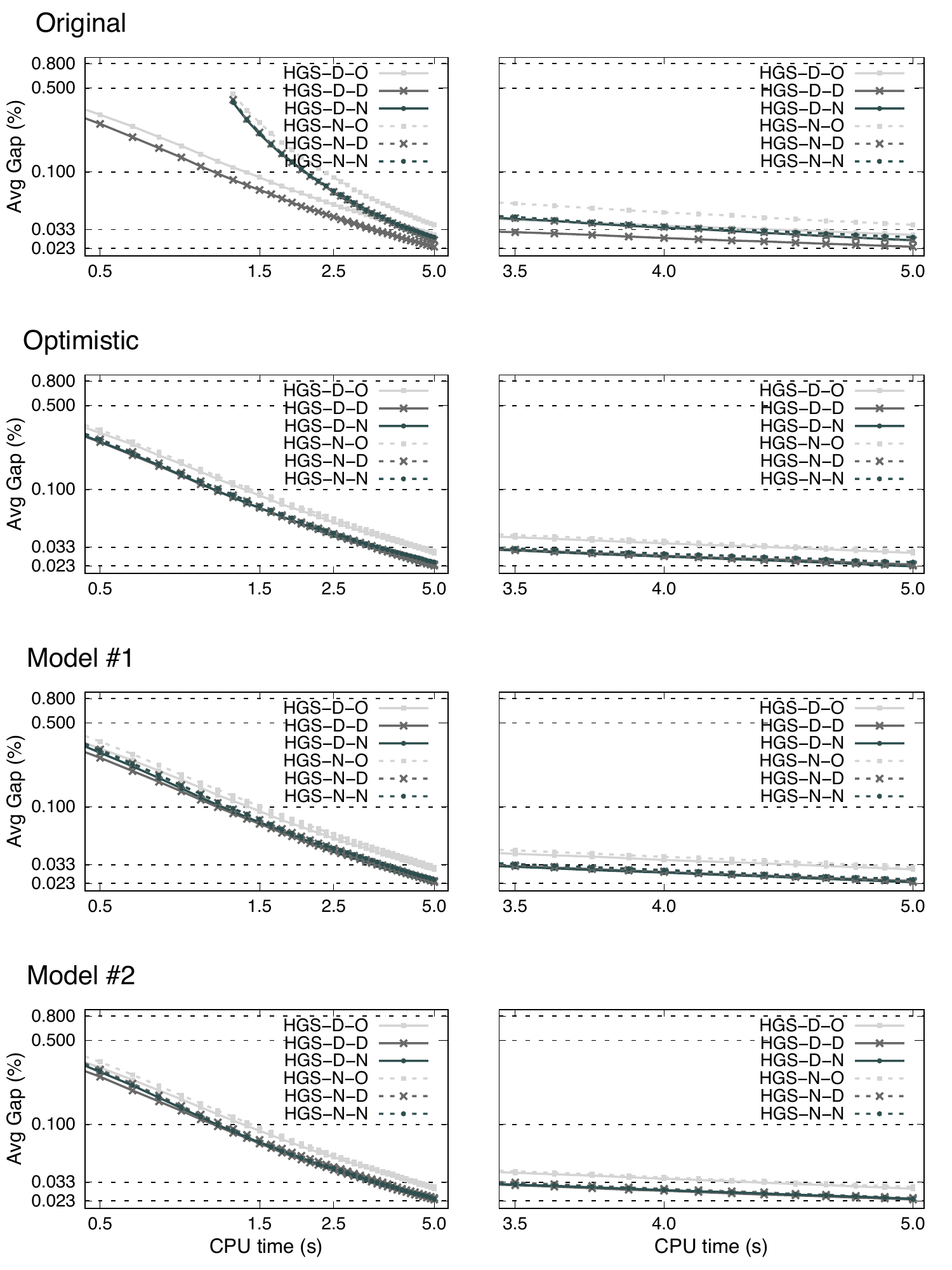}
		\caption{Convergence plots on set~XML for all GNN configurations (for each configuration, left graph = complete run, right graph = last 1.5\,seconds of the run time)}
		\label{fig:convergence-plot-XML-appendix}
	\end{center}
\end{figure*}

Additionally, Figure~\ref{fig:convergence-plot-XML-appendix} provides convergence plots for all possible GNN configurations.
The graphs included in it represent the progress of the average gap over time, considering the parameter value $\Gamma=15$. 
On these graphs, it is noteworthy that the \textsc{Original} configuration of the GNN requires significant inference time to generate the heatmap before the optimization can start. In comparison, \textsc{Model \#1} and \textsc{Model \#2} start much faster and behave quite similarly to the \textsc{Optimistic} curve that ignores inference time.

\begin{figure*}[hp!]
	
	\begin{subfigure}{.5\textwidth}
		\centering
		\includegraphics[width=\textwidth]{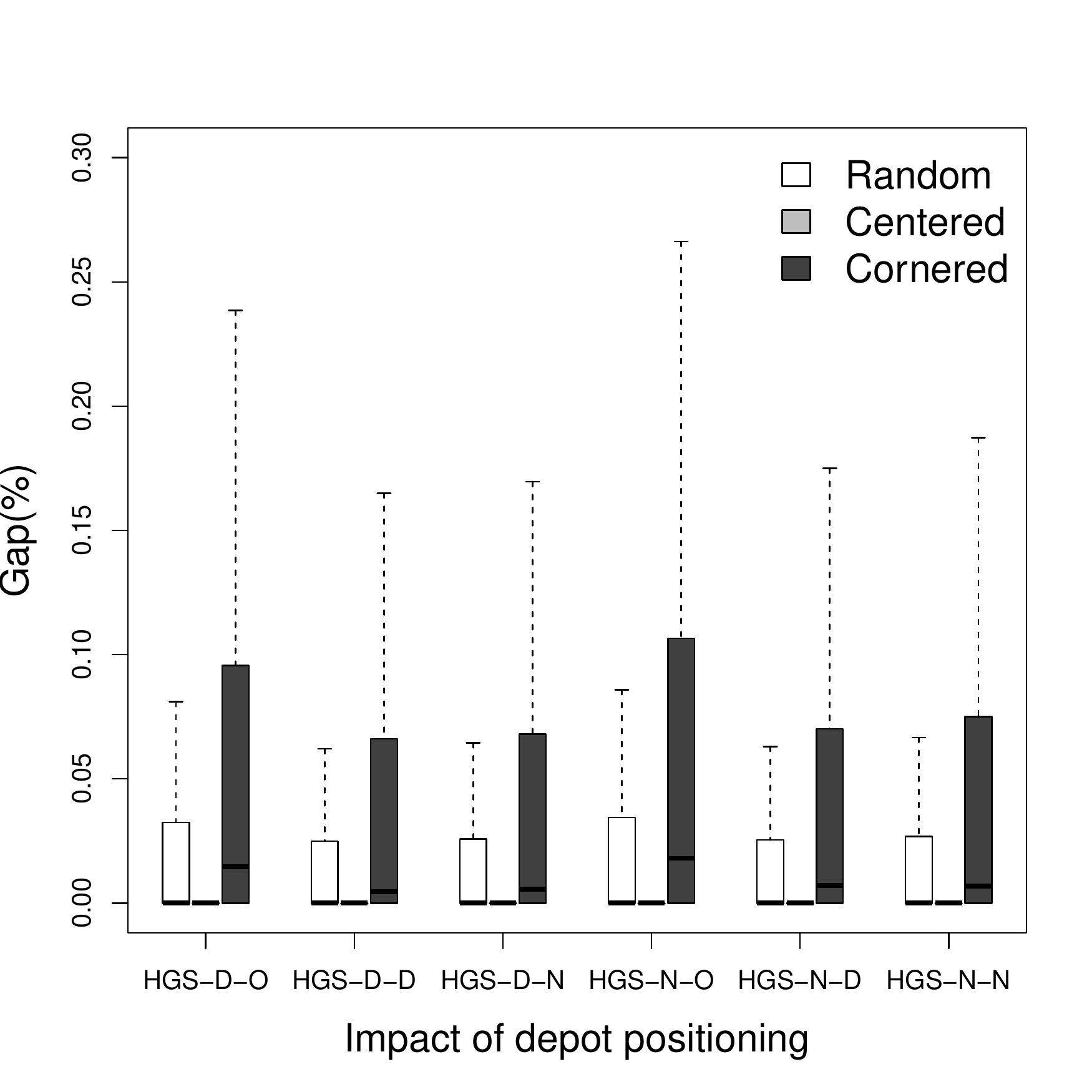}
	\end{subfigure}%
	\begin{subfigure}{.5\textwidth}
		\centering
		\includegraphics[width=1\textwidth]{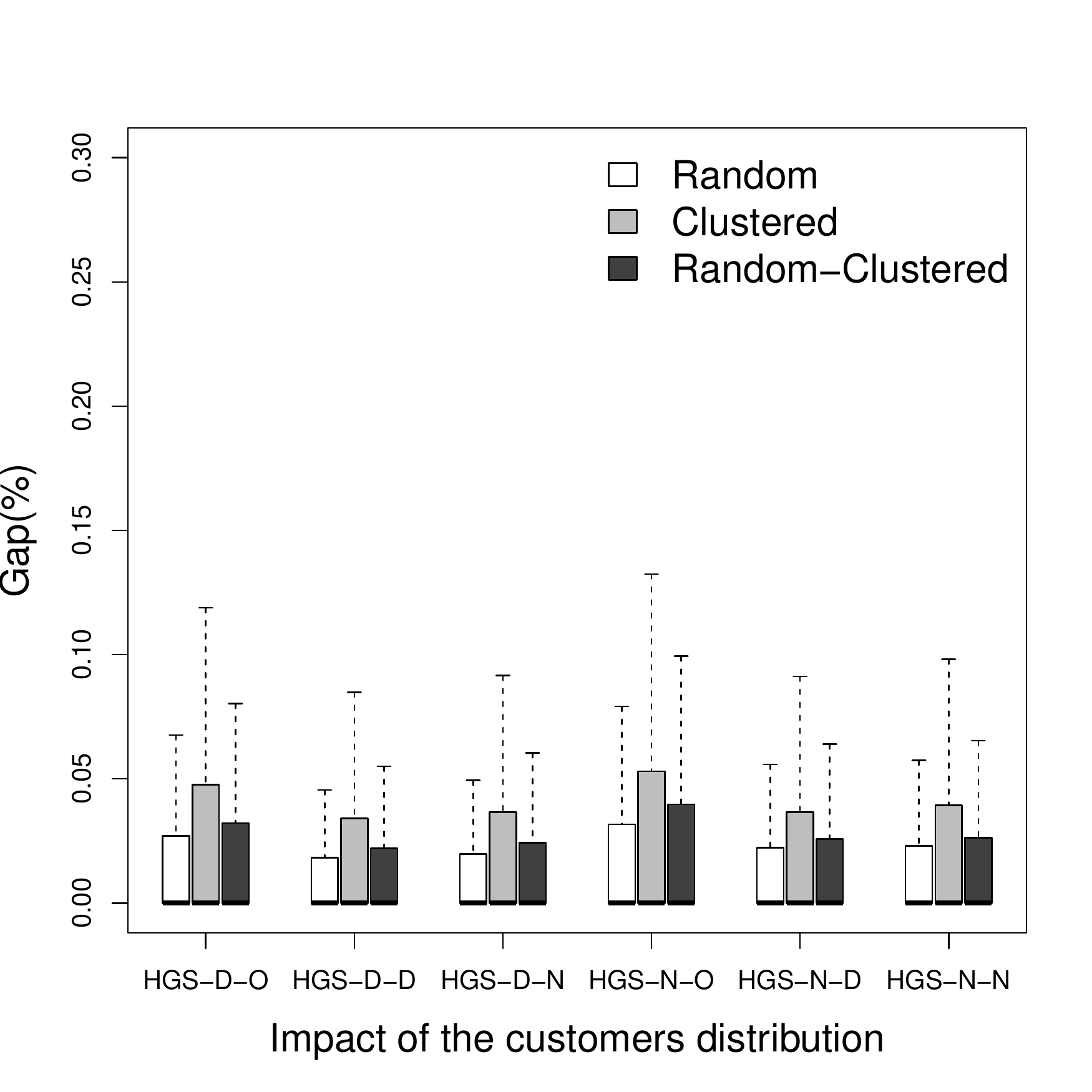}
	\end{subfigure}
	\begin{subfigure}{.5\textwidth}
		\centering
		\includegraphics[width=\textwidth]{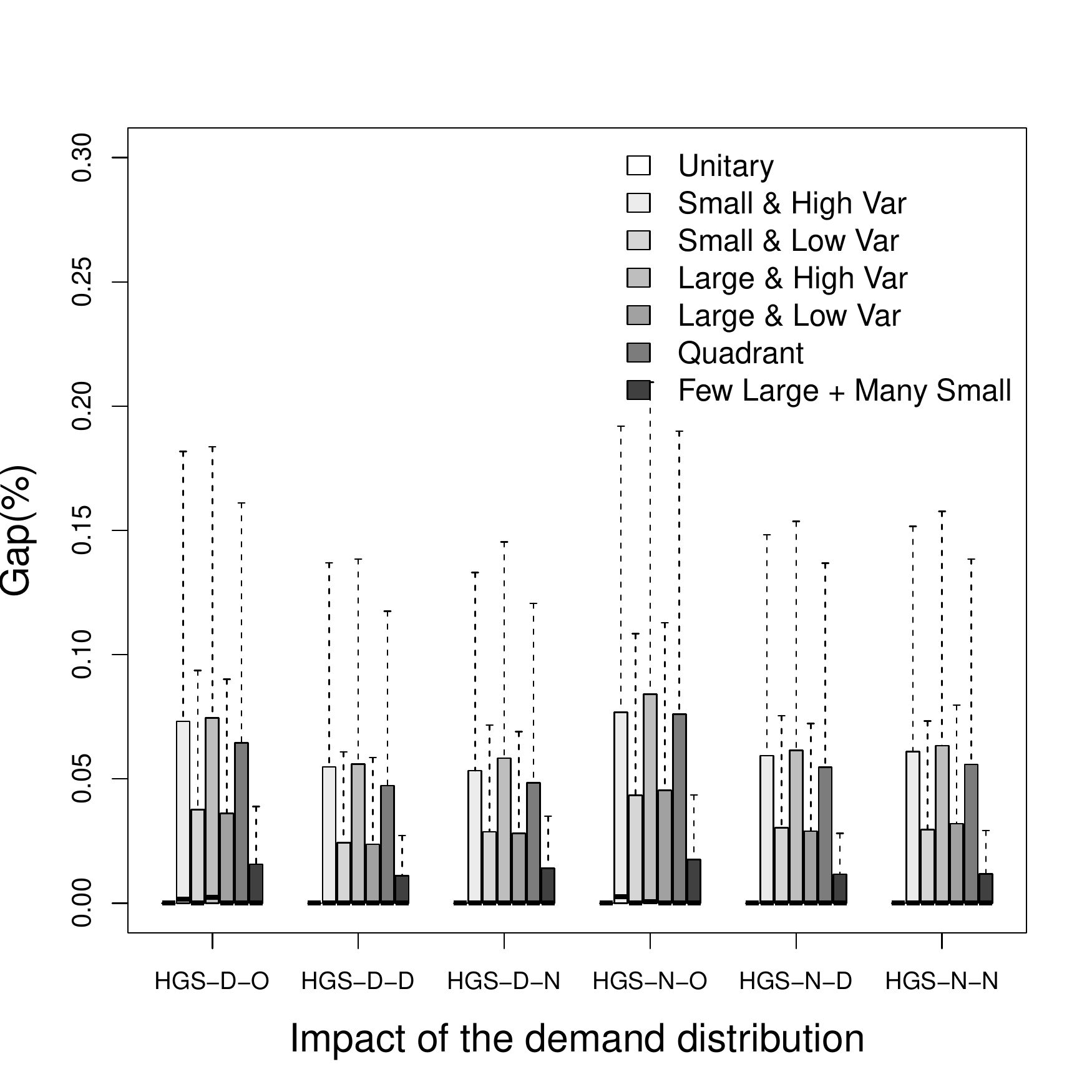}
	\end{subfigure}%
	\begin{subfigure}{.5\textwidth}
		\centering
		\includegraphics[width=\textwidth]{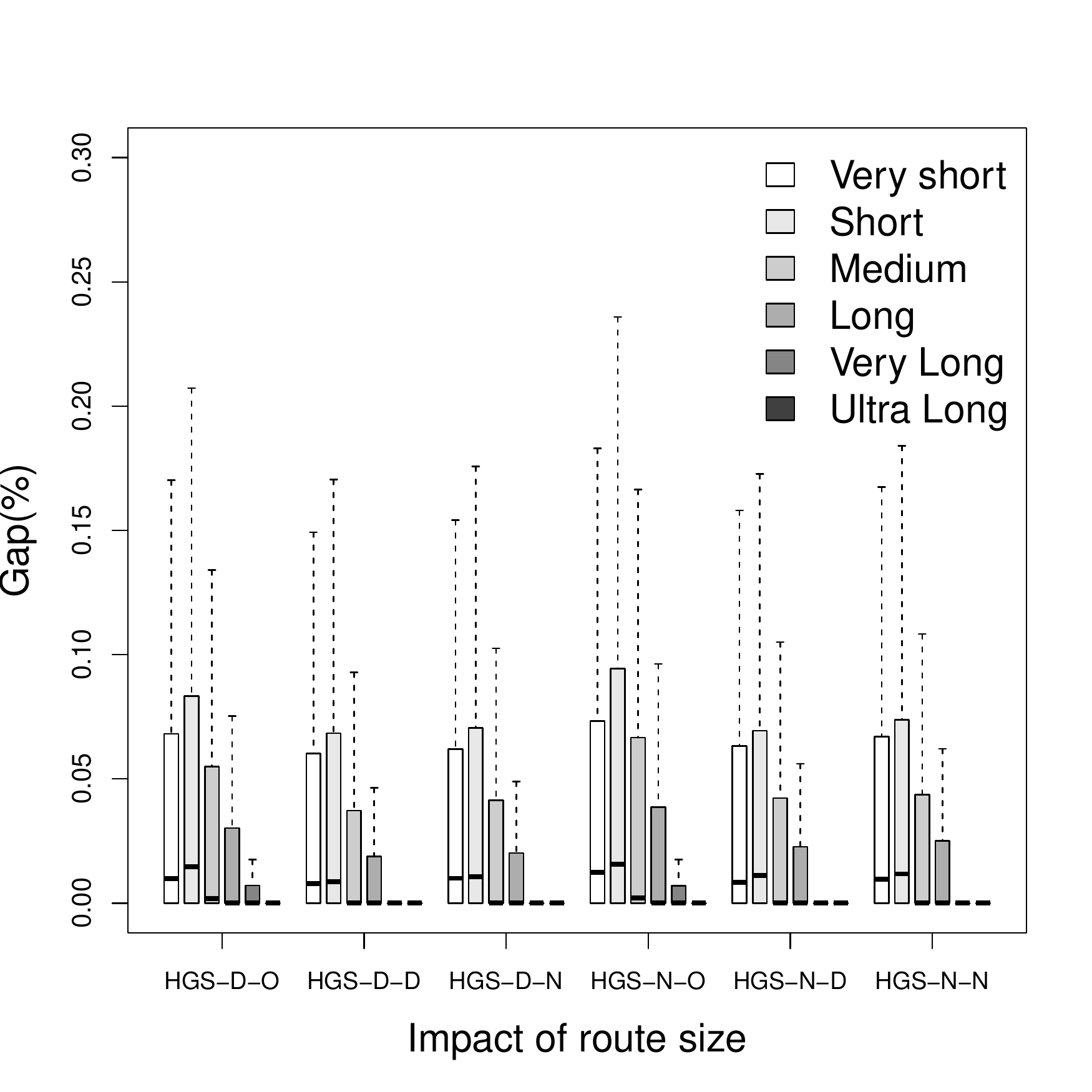}
	\end{subfigure}
	\caption{Boxplots of the percentage error gaps achieved by the methods, for different subsets of the XML~instances, using $\Gamma = 15$ and the \textsc{Original} configuration of the GNN}
	\label{fig:boxplot-XML-appendix}
\end{figure*}

Next, Figure~\ref{fig:boxplot-XML-appendix} presents boxplots of the percentage error gaps of the methods for different subsets of the XML instances, using the \textsc{Original} configuration of the GNN and $\Gamma =15$. Indeed, as discussed in \citet{Uchoa2017} and \citet{Queiroga2022}, the instances were designed to mimic important features present in real-world problems, with different conventions regarding depot position, customers position, demand distribution, and average route size. Therefore, each boxplot corresponds to a subset of the instances with a specific configuration for one of these parameters. In these boxplots, the boxes indicate the first and third quartile, and the whiskers extend to $1.5$ times the interquartile range.

\begin{table*}[hp!]
	\caption{Results for set~XML for all values of $\Gamma \in \{15,20,30,50\}$}
	\label{tab:results-XML-appendix}
	\centering
	\begin{small}
		\begin{sc}
			\resizebox{0.93\textwidth}{!}
			{
				\begin{tabular}{crrrrrrrrrrrr}
					\toprule
					& \multicolumn{2}{c}{HGS-D-O} & \multicolumn{2}{c}{HGS-D-D} & \multicolumn{2}{c}{HGS-D-N} & \multicolumn{2}{c}{HGS-N-O}  & \multicolumn{2}{c}{HGS-N-D}  & \multicolumn{2}{c}{HGS-N-N} \\
					GNN & \#Opt &  Gap\% &  \#Opt &  Gap\% &  \#Opt &  Gap\% &  \#Opt &  Gap\%&  \#Opt &  Gap\%&  \#Opt &  Gap\%\\
					\midrule
					\\
					\multicolumn{13}{l}{Original (\#Nodes: 300; \#Hidden Layers: 30; \#Epochs: 1500; Pred-T(s) = 0.85)}\\
					15 &7715 &0.030 &\textbf{8105} &0.024 &8086 &\textbf{0.023} &7691 &0.031 &8046 &\textbf{0.023} &8011 & 0.025\\
					20 &7612 &0.032 &\textbf{7969} &\textbf{0.025} &7932 &0.026 &7482 &0.035 &7903 &0.027 &7811 & 0.028\\
					30 &7181 &0.041 &\textbf{7590} &\textbf{0.031} &7544 &0.032 &7054 &0.044 &7466 &0.034 &7419 & 0.036\\
					50 &6421 &0.063 &\textbf{6804} &\textbf{0.050} &6775 &0.052 &6365 &0.067 &6674 &0.055 &6684 & 0.055\\
					\midrule
					Avg  &7232.3 &0.042 &\textbf{7617.0} & \textbf{0.033} &7584.3 &\textbf{0.033} &7148.0 &0.044 &7522.3 &0.035 &7481.3 & 0.036\\
					\midrule
					\\
					\multicolumn{13}{l}{Optimistic (\#Nodes: 300; \#Hidden Layers: 30; \#Epochs: 1500; Pred-T(s) = Ignored)}\\
					15 &7715 &0.030 &8105 &0.024 &\textbf{8120} &\textbf{0.023} &7732 &0.031 &8062 &\textbf{0.023} &8041 & 0.025\\
					20 &7612 &0.032 &\textbf{7969} &\textbf{0.025} &7956 &0.026 &7515 &0.034 &7942 &0.026 &7814 & 0.028\\
					30 &7181 &0.041 &7590 &\textbf{0.031} &\textbf{7591} &\textbf{0.031} &7102 &0.044 &7517 &0.033 &7465 & 0.035\\
					50 &6421 &0.063 &6804 &\textbf{0.050} &\textbf{6830} &0.052 &6427 &0.066 &6718 &0.054 &6748 & 0.054\\\midrule
					Avg  &7232.3 &0.042 &7617.0 &\textbf{0.033} &\textbf{7624.3} &\textbf{0.033} &7194.0 &0.044 &7559.8 &0.034 &7517.0 & 0.036\\
					\midrule
					\\
					\multicolumn{10}{l}{\textsc{Model \#1} (\#Nodes: 10; \#Hidden Layers: 5; \#Epochs: 500; Pred-T(s) = 0.03)}\\
					15 &7715 &0.030 &\textbf{8105} &0.024 &8094 &\textbf{0.023} &7697 &0.031 &8028 &0.024 &7994 & 0.025 \\
					20 &7612 &0.032 &\textbf{7969} &\textbf{0.025} &7941 &\textbf{0.025} &7450 &0.034 &7822 &0.027 &7865 & 0.027 \\
					30 &7181 &0.041 &\textbf{7590} &\textbf{0.031} &7524 &0.032 &7068 &0.044 &7464 &0.034 &7384 & 0.036 \\
					50 &6421 &0.063 &6804 &\textbf{0.050} &\textbf{6815} &0.051 &6379 &0.066 &6709 &0.056 &6715 & 0.055 \\\midrule
					Avg  &7232.3 &0.042 &\textbf{7617.0} &\textbf{0.033} &7593.5 &\textbf{0.033}&7148.5 &0.044 &7505.8 &0.035 &7489.5 & 0.036 \\
					\midrule
					\\
					\multicolumn{13}{l}{\textsc{Model \#2} (\#Nodes: 10; \#Hidden Layers: 5; \#Epochs: 1500; Pred-T(s) = 0.03)}\\
					15 &7715 &0.030 &\textbf{8105} &\textbf{0.024} &8102 &\textbf{0.024} &7719 &0.030 &8067 &\textbf{0.024} &8057 & \textbf{0.024} \\
					20 &7612 &0.032 &\textbf{7969} &\textbf{0.025} &7954 &\textbf{0.025} &7577 &0.033 &7845 &0.027 &7900 & 0.026 \\
					30 &7181 &0.041 &\textbf{7590} &\textbf{0.031} &7582 &\textbf{0.031} &7105 &0.042 &7545 &0.033 &7503 & 0.033 \\
					50 &6421 &0.063 &6804 &0.050 &\textbf{6830} &\textbf{0.049} &6394 &0.065 &6724 &0.054 &6758 & 0.053 \\\midrule
					Avg  &7232.3 &0.042 &\textbf{7617.0} &0.033 &\textbf{7617.0} &\textbf{0.032}&7198.8 &0.043 &7545.3 &0.035 &7554.5 & 0.034 \\
					\bottomrule
				\end{tabular}
			}
		\end{sc}
	\end{small}
\end{table*}

\section{Detailed Experimental Results -- Set X}

\begin{figure*}[hp!]
	\begin{center}
		\includegraphics[width=0.7\textwidth]{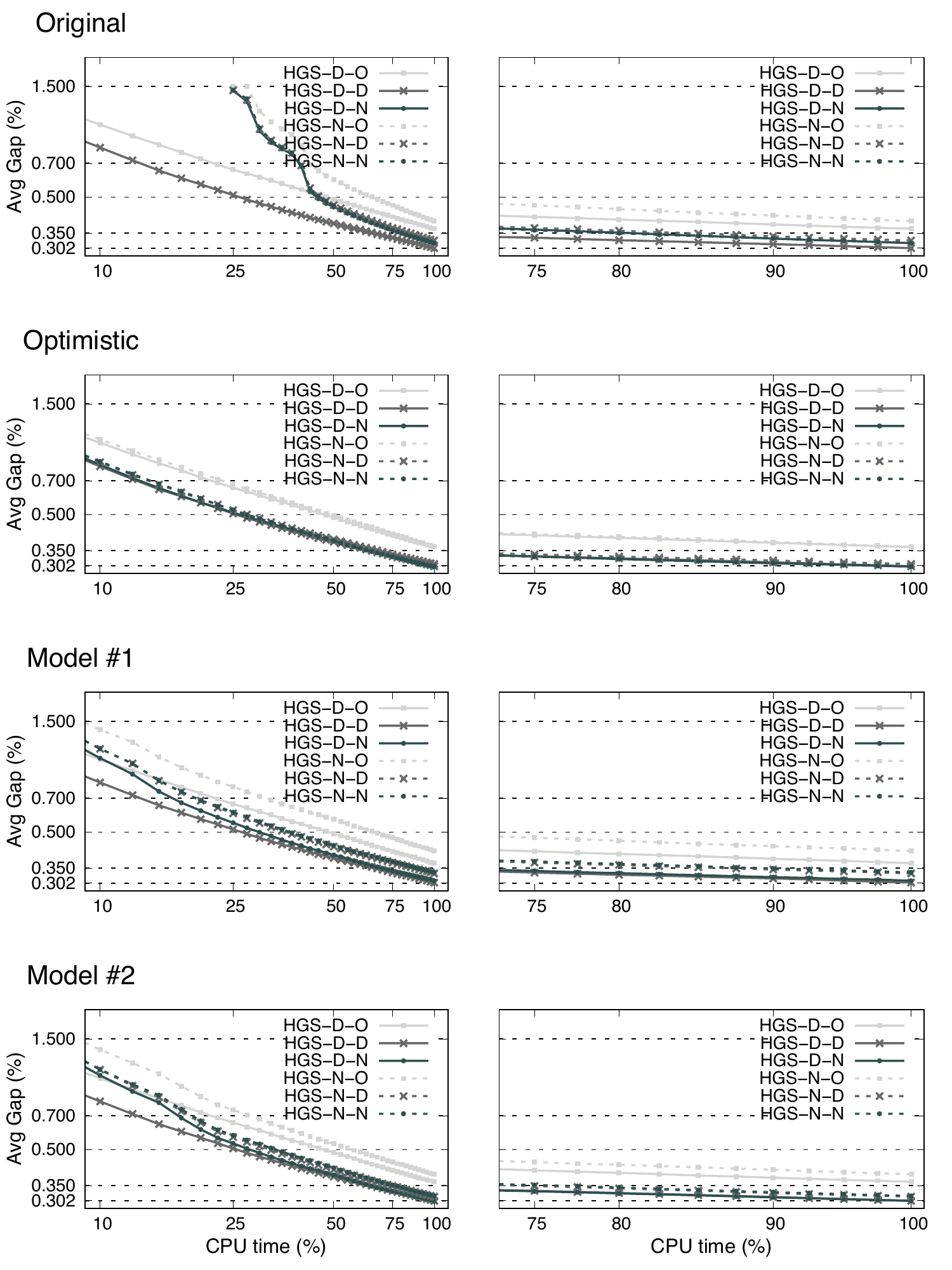}
		\caption{Convergence plots on set~X for all GNN configurations (for each configuration, left graph = complete run, right graph = last 25~\% of the run time)}
		\label{fig:convergence-plot-X-appendix}
	\end{center}
\end{figure*}

\begin{figure*}[hp!]
	\begin{subfigure}{.5\textwidth}
		\centering
		\includegraphics[width=\textwidth]{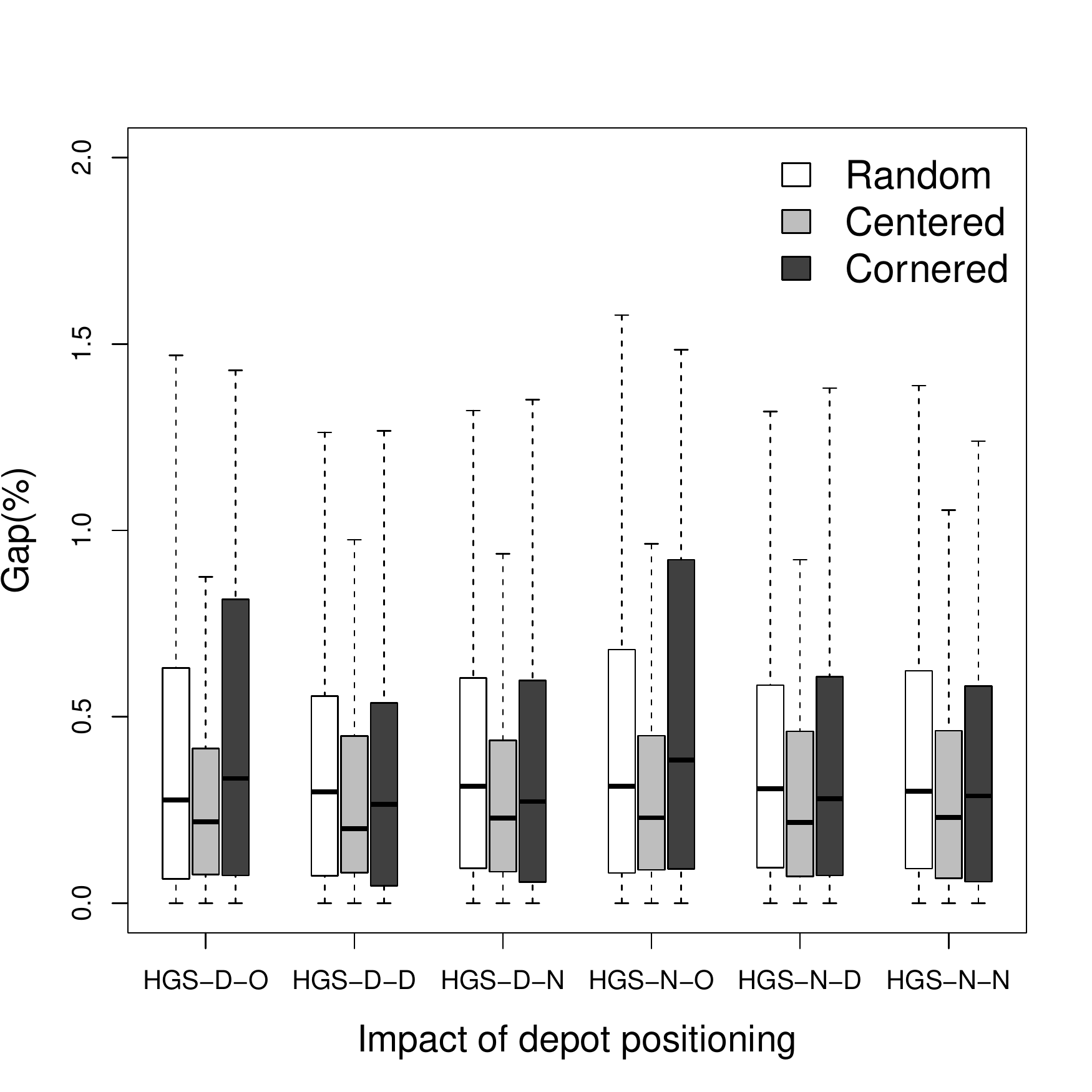}
		\label{fig:box-plot-depot-X}
	\end{subfigure}%
	\begin{subfigure}{.5\textwidth}
		\centering
		\includegraphics[width=1\textwidth]{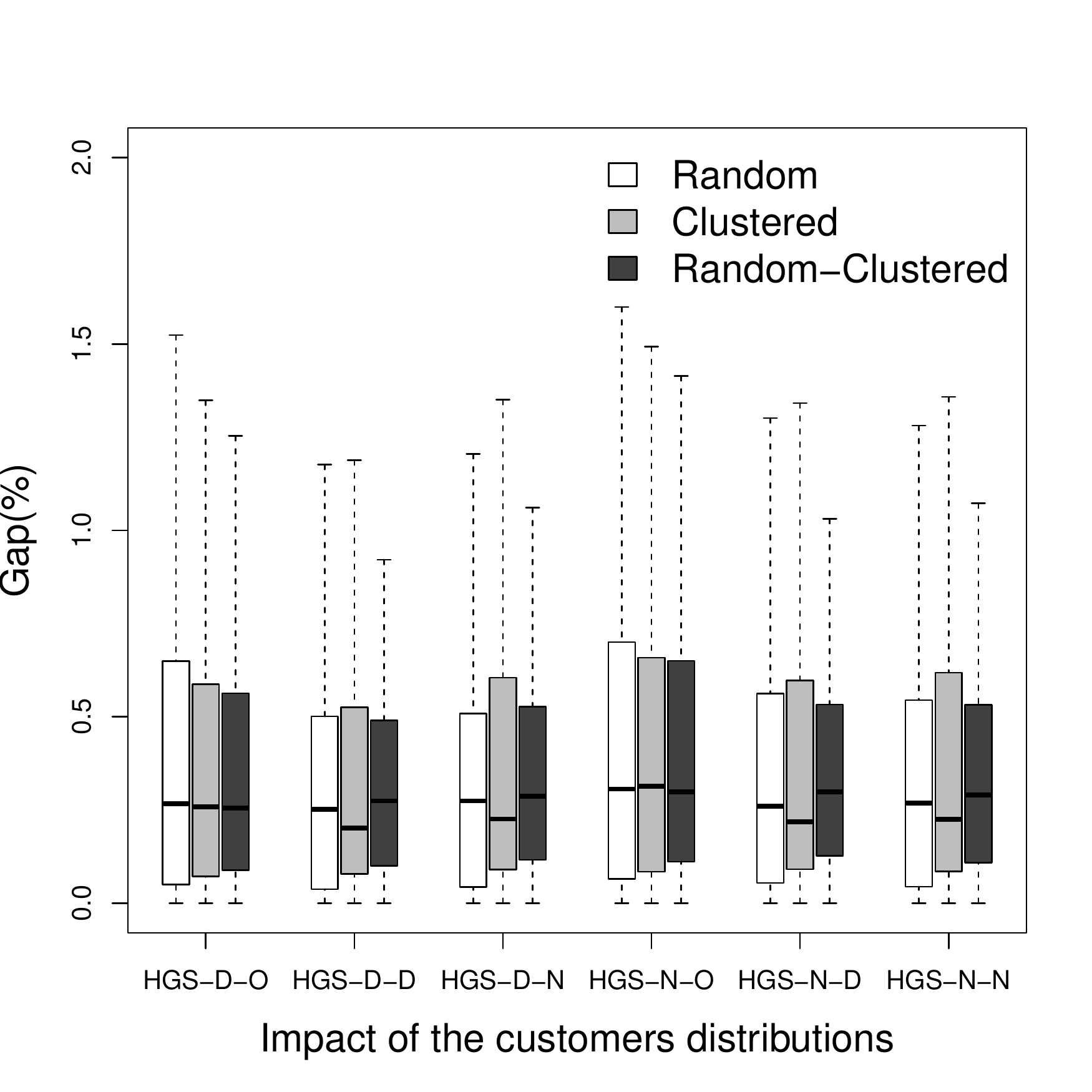}
		\label{fig:box-plot-customers-X}
	\end{subfigure}
	\begin{subfigure}{.5\textwidth}
		\centering
		\includegraphics[width=\textwidth]{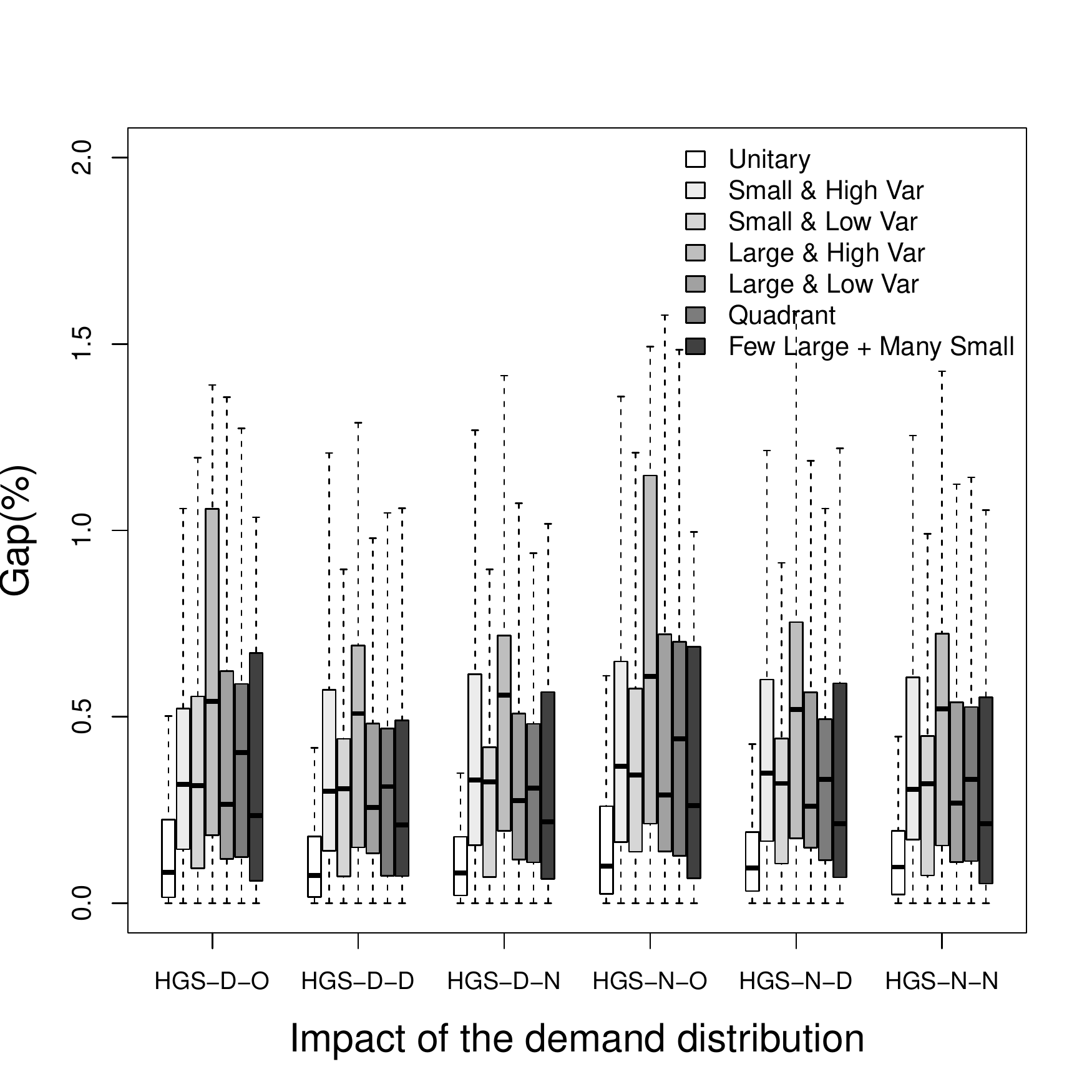}
		\label{fig:box-plot-demand-X}
	\end{subfigure}%
	\begin{subfigure}{.5\textwidth}
		\centering
		\includegraphics[width=\textwidth]{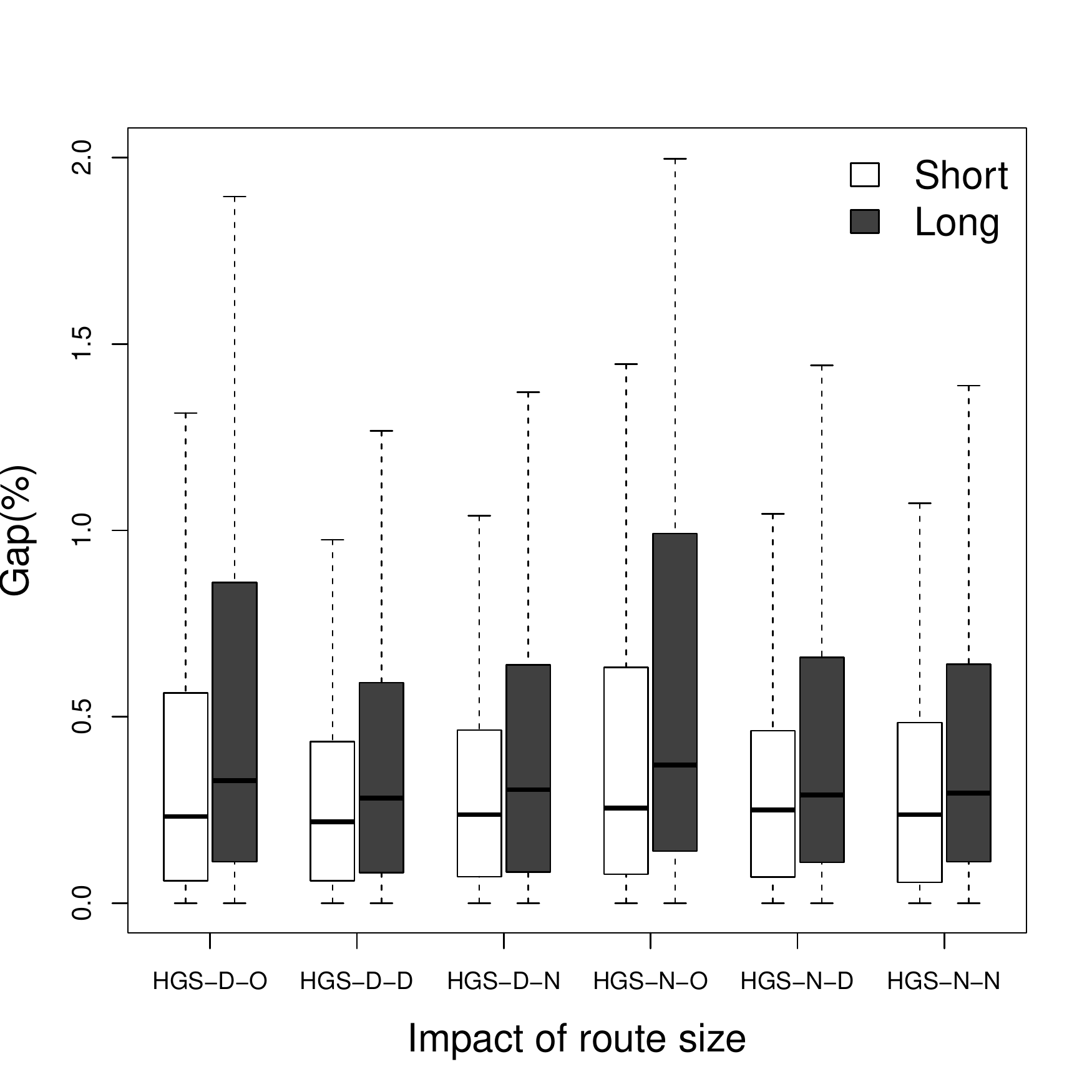}
		\label{fig:box-plot-routesize-X}
	\end{subfigure}
	\caption{Boxplots of the percentage error gaps achieved by the methods, for different subsets of the X~instances, using $\Gamma = 15$ and the \textsc{Original} configuration of the GNN}
	\label{fig:boxplot-X-appendix}
\end{figure*}

Finally, Table~\ref{tab:results-X-appendix} and Figures~\ref{fig:convergence-plot-X-appendix}--\ref{fig:boxplot-X-appendix} provide the same detailed results for the X instance set of \cite{Uchoa2017}.

\begin{table*}[hp!]
	\caption{Results for set~X for all values of $\Gamma \in \{15,20,30,50\}$}
	\label{tab:results-X-appendix}
	\centering
	\begin{small}
		\begin{sc}
			\resizebox{0.93\textwidth}{!}
			{
				\begin{tabular}{crrrrrrrrrrrr}
					\toprule
					&\multicolumn{2}{c}{HGS-D-O} &  \multicolumn{2}{c}{HGS-D-D} & \multicolumn{2}{c}{HGS-D-N}  &  \multicolumn{2}{c}{HGS-N-O}  & \multicolumn{2}{c}{HGS-N-D}&  \multicolumn{2}{c}{HGS-N-N} \\
					GNN &  \#Opt &   Gap\% &   \#Opt &   Gap\% &   \#Opt &   Gap\%&   \#Opt &   Gap\%&   \#Opt &   Gap\% &   \#Opt &   Gap\%\\
					\midrule
					\\
					\multicolumn{13}{l}{Original (\#Nodes: 300; \#Hidden Layers: 100; \#Epochs: 1500; Pred-T(s) = 15.2\%)}\\
					15& \textbf{188}& 0.368& 187& \textbf{0.302}& 184& 0.317& 177& 0.395& 169& 0.325& 172&  0.317\\
					20& 166& 0.380& \textbf{186}& \textbf{0.305}& 184& 0.323& 160& 0.412& 169& 0.328& 175&  0.322\\
					30& 162& 0.401& \textbf{176}& \textbf{0.318}& 164& 0.341& 139& 0.438& 173& 0.355& 160&  0.348\\
					50& 130& 0.464& \textbf{158}& \textbf{0.370}& 140& 0.402& 114& 0.522& 146& 0.410& 134&  0.411\\\midrule
					Avg & 161.5& 0.403& \textbf{176.8}& \textbf{0.324}& 168.0& 0.346& 147.5& 0.442& 164.3& 0.354& 160.3&  0.350\\
					\midrule
					\\
					\multicolumn{13}{l}{Optimistic (\#Nodes: 300; \#Hidden Layers: 100; \#Epochs: 1500; Pred-T(s) = \textsc{Ignored})}\\
					15& \textbf{188}& 0.368& 187& 0.302&187& \textbf{0.299}& 185& 0.365& 177& 0.307& 182&  \textbf{0.299}\\
					20& 166& 0.380& 186& \textbf{0.305}& \textbf{191}& 0.306& 170& 0.383& 176& 0.308& 179& \textbf{0.305}\\
					30& 162& 0.401& \textbf{176}& \textbf{0.318}&  171& 0.320& 153& 0.406& \textbf{176}& 0.331& 167&  0.326\\
					50& 130& 0.464& \textbf{158}& \textbf{0.370}& 153& 0.372& 122& 0.482& 154& 0.380& 142&  0.382\\\midrule
					Avg & 161.5& 0.403& \textbf{176.8}& \textbf{0.324}& 175.5& \textbf{0.324}& 157.5& 0.409& 170.8& 0.332& 167.5&  0.328\\
					\midrule
					\\
					\multicolumn{13}{l}{\textsc{Model \#1} (\#Nodes: 10; \#Hidden Layers: 5; \#Epochs: 500; Pred-T(s) = 1.9\%)}\\
					15& \textbf{188}& 0.368& 187& \textbf{0.302}&185& 0.309& 166& 0.414& 177& 0.332& 184&  0.336\\
					20& 166& 0.380& \textbf{186}& 0.305& 172& 0.312& 165& 0.419& 177& 0.337& 179&  0.340\\
					30& 162& 0.401& 176& 0.318& \textbf{181}& 0.322& 149& 0.445& 170& 0.355& 164&  0.350\\
					50& 130& 0.464& \textbf{158}& 0.370& 143& 0.373& 120& 0.500& 149& 0.391& 143&  0.400\\\midrule
					Avg & 161.5& 0.403& \textbf{176.8}& \textbf{0.324}& 170.3& 0.329& 150.0& 0.445& 168.3& 0.354& 167.5&  0.357\\
					\midrule
					\\
					\multicolumn{13}{l}{\textsc{Model \#2} (\#Nodes: 10; \#Hidden Layers: 5; \#Epochs: 1500; Pred-T(s) = 1.9\%)}\\
					15& \textbf{188}& 0.368& 187& 0.302& 186& \textbf{0.301}& 169& 0.391& 175& 0.314& 170&  0.316\\
					20& 166& 0.380& \textbf{186}& \textbf{0.305}& 179& 0.310& 162& 0.394& 183& 0.321& 177&  0.322\\
					30& 162& 0.401& \textbf{176}& \textbf{0.318}&  174& 0.322& 164& 0.429& 162& 0.335& 174&  0.340\\
					50& 130& 0.464& \textbf{158}& \textbf{0.370}& 155& 0.371& 127& 0.478& 150& 0.388& 146&  0.389\\\midrule
					Avg & 161.5& 0.403& \textbf{176.8}& \textbf{0.324}& 173.5& 0.326& 155.5& 0.423& 167.5& 0.340& 166.8&  0.342\\
					\bottomrule
				\end{tabular}
			}
		\end{sc}
	\end{small}
\end{table*} 

\end{document}